\documentclass[a4paper,fleqn]{ITSHA}

\usepackage[numbers]{natbib}
\usepackage{amsmath}
\usepackage{algorithm}
\usepackage{algorithmic}
\usepackage{caption}
\usepackage{subfigure}
\usepackage{multirow}
\usepackage{booktabs}
\usepackage{eqparbox}

\def\tsc#1{\csdef{#1}{\textsc{\lowercase{#1}}\xspace}}
\tsc{WGM}
\tsc{QE}
\tsc{EP}
\tsc{PMS}
\tsc{BEC}
\tsc{DE}
\begin{document}
\let\WriteBookmarks\relax
\def\floatpagepagefraction{1}
\def\textpagefraction{.001}
\shorttitle{An Effective Iterated Two-stage Heuristic Algorithm for the $m$TSP}
\shortauthors{Zheng et~al.}

\title [mode = title]{An Effective Iterated Two-stage Heuristic Algorithm for the Multiple Traveling
Salesmen Problem}

\author[1]{Jiongzhi Zheng}%[type=editor,
                        %auid=000,bioid=1,
                        %prefix=Sir,
                        %role=Researcher,
                        %orcid=0000-0001-7511-2910]
%\ead{jzzheng@hust.edu.cn}
%\ead[url]{www.cvr.cc, cvr@sayahna.org}

\author[1]{Yawei Hong}
\author[1]{Wenchang Xu}
\author[1]{Wentao Li}

\credit{Conceptualization of this study, Methodology, Software, Writing and revision}

%\address[1]{School of Computer Science, Huazhong University of Science and Technology, China}
%\address[2]{Institute of Artificial Intelligence, Huazhong University of Science and Technology, China}
\address[1]{School of Mechanical Science and Engineering, Huazhong University of Science and Technology, China}

\author[1]{Yongfu Chen}%[role=Supervisor]
\cormark[1]
%\ead{chenyf@hust.edu.cn}
\cortext[cor1]{Corresponding author}
\credit{Conceptualization of this study, Methodology, Supervision, Writing and revision}

\iffalse
\author[1,3]{CV Radhakrishnan}[type=editor,
                        auid=000,bioid=1,
                        prefix=Sir,
                        role=Researcher,
                        orcid=0000-0001-7511-2910]
\cormark[1]
\fnmark[1]
\ead{cvr_1@tug.org.in}
\ead[url]{www.cvr.cc, cvr@sayahna.org}

\credit{Conceptualization of this study, Methodology, Software}

\address[1]{Elsevier B.V., Radarweg 29, 1043 NX Amsterdam, The Netherlands}

\author[2,4]{Han Theh Thanh}[style=chinese]

\author[2,3]{CV Rajagopal}[%
   role=Co-ordinator,
   suffix=Jr,
   ]
\fnmark[2]
\ead{cvr3@sayahna.org}
\ead[URL]{www.sayahna.org}

\credit{Data curation, Writing - Original draft preparation}

\address[2]{Sayahna Foundation, Jagathy, Trivandrum 695014, India}

\author%
[1,3]
{Rishi T.}
\cormark[2]
\fnmark[1,3]
\ead{rishi@stmdocs.in}
\ead[URL]{www.stmdocs.in}

\address[3]{STM Document Engineering Pvt Ltd., Mepukada,
    Malayinkil, Trivandrum 695571, India}

\cortext[cor1]{Corresponding author}
\cortext[cor2]{Principal corresponding author}
\fntext[fn1]{This is the first author footnote. but is common to third
  author as well.}
\fntext[fn2]{Another author footnote, this is a very long footnote and
  it should be a really long footnote. But this footnote is not yet
  sufficiently long enough to make two lines of footnote text.}

\nonumnote{This note has no numbers. In this work we demonstrate $a_b$
  the formation Y\_1 of a new type of polariton on the interface
  between a cuprous oxide slab and a polystyrene micro-sphere placed
  on the slab.
  }

\fi

\begin{abstract}
The multiple Traveling Salesmen Problem ($m$TSP) is a general extension of the famous NP-hard Traveling Salesmen Problem (TSP), that there are $m$ ($m>1$) salesmen to visit the cities. In this paper, we address the $m$TSP with both the \emph{minsum} objective and \emph{minmax} objective, which aims at minimizing the total length of the $m$ tours and the length of the longest tour among all the $m$ tours, respectively. We propose an iterated two-stage heuristic algorithm called ITSHA for the $m$TSP. Each iteration of ITSHA consists of an initialization stage and an improvement stage. The initialization stage aims to generate high-quality and diverse initial solutions. The improvement stage mainly applies the variable neighborhood search (VNS) approach based on our proposed effective local search neighborhoods to optimize the initial solution. Moreover, some local optima escaping approaches are employed to enhance the search ability of the algorithm. Extensive experimental results on a wide range of public benchmark instances show that ITSHA significantly outperforms the state-of-the-art heuristic algorithms in solving the $m$TSP on both the objectives.

%\noindent\texttt{\textbackslash begin{abstract}} \dots 
%\texttt{\textbackslash end{abstract}} and
%\verb+\begin{keyword}+ \verb+...+ \verb+\end{keyword}+ 
%which
%contain the abstract and keywords respectively. 

%\noindent Each keyword shall be separated by a \verb+\sep+ command.
\end{abstract}

%\includegraphics{grabs.pdf}
\iffalse
\begin{highlights}
\item We propose an effective ITSHA local search for the \emph{minsum} and \emph{minmax} $m$TSP.
%\item We address the NP-hard $m$TSP with \emph{minsum} and \emph{minmax} objectives.
%\item We propose an iterated two-stage heuristic algorithm for solving the problem.
\item We propose three effective local search neighborhoods for the $m$TSP.
\item We propose several approaches for escaping from the local optima to improve ITSHA.
\item Experimental results demonstrate the superiority of the proposed algorithm.
\end{highlights}
\fi


\maketitle

\section{Introduction}
The multiple Traveling Salesmen Problem ($m$TSP) is a general extension of the famous NP-hard Traveling Salesmen Problem (TSP), that there are $m$ ($m>1$) salesmen to visit the cities. In this paper, we consider the $m$TSP with a single depot, that is, the $m$ salesmen travel from the same depot to visit all the cities exactly once without overlapping and finally return to the depot. We address the $m$TSP with both the \emph{minsum} and \emph{minmax} objectives, which aims at minimizing the total length of the $m$ tours and the length of the longest tour among all the $m$ tours, respectively. The $m$TSP is not only a natural but also a more practical extension of the TSP, that finds many practical applications in the real world \cite{Cheikhrouhou2021}. For example, the well-known Vehicle Routing Problem (VRP) \cite{Nagata2010,Arnold2019}, production schedules \cite{Tang2000}, the school bus routing problem \cite{Miranda2018}, printing press schedules \cite{Carter2002}, task allocation \cite{Vandermeulen2019}, etc.

Typical methods for the $m$TSP are mainly exact algorithms \cite{Franca1995,Kara2006}, approximation algorithms \cite{Frederickson1978}, and heuristics \cite{Venkatesh2015,Soylu2015,Lu2019}. The exact algorithms may be difficult for large instances and the approximation algorithms may suffer from weak optimality guarantees. Heuristics are known to be the most efficient and effective approaches for solving the $m$TSP. %For representative and effective heuristic algorithms, we have the artificial bee colony (ABC) algorithms \cite{Venkatesh2015}, the invasive weed optimization (IWO) algorithm \cite{Venkatesh2015}, the general variable neighborhood search (GVNS) heuristic \cite{Soylu2015}, the ant colony optimization (ACO) algorithm \cite{Lu2019} and the evolution strategy (ES) approach \cite{Karabulut2021}. These heuristics can be used to solve both the \emph{minsum} and \emph{minmax} $m$TSP.

Population-based meta-heuristics are the most popular and effective heuristic algorithms for the \emph{minsum} and \emph{minmax} $m$TSP recently. Some of them address both of the two objectives of $m$TSP. For example, the genetic algorithms (GA) \cite{Carter2006,Singh2009,Yuan2013}, artificial bee colony (ABC) algorithms \cite{Venkatesh2015}, ant
colony optimization (ACO) algorithms \cite{Lu2019,Liu2009}, evolution strategy (ES) algorithm \cite{Karabulut2021}, and other population-based approaches \cite{Venkatesh2015}. Among these algorithms, the ABC algorithms as well as the invasive weed optimization (IWO) algorithm proposed by Venkatesh and Singh \cite{Venkatesh2015}, the ACO algorithm \cite{Lu2019}, and the ES algorithm \cite{Karabulut2021} are some of the state-of-the-art heuristics for both the \emph{minsum} and \emph{minmax} $m$TSP. In addition, some studies focus on one of the two objectives of $m$TSP. The genetic algorithm called GAL \cite{Lo2018} and the memetic algorithm called MASVND \cite{Wang2017} are two effective heuristics for the \emph{minsum} and \emph{minmax} $m$TSP, respectively. They tested their algorithms on the $m$TSP instances with more than one thousand cities.

Local search is another type of heuristic algorithm, which is widely used in some famous combinatorial optimization problems such as the TSP \cite{Lin1973,Helsgaun2000}, satisfiability \cite{Selman1992}, and maximum satisfiability \cite{Luo2017}. However, the local search technique is mainly applied to improve the population-based $m$TSP meta-heuristics by incorporating with them in the related studies \cite{Venkatesh2015,Karabulut2021,Wang2017}, and few researches employ the local search method to directly solve the standard \emph{minsum} and \emph{minmax} $m$TSP. Among the local search algorithms, the general variable neighborhood search (GVNS) algorithm proposed by Soylu \cite{Soylu2015} is one of the best-performing based on variable neighborhood search (VNS) for the \emph{minsum} and \emph{minmax} $m$TSP. 

This work aims at making up the lack of effective local search heuristics for solving the \emph{minsum} and \emph{minmax} $m$TSP by introducing an iterated two-stage heuristic algorithm, denoted as ITSHA. %In the ITSHA algorithm, we define several much more effective and efficient neighborhoods than the neighborhoods used in previous algorithms \cite{Soylu2015,Wang2017}. We apply the . 
The first stage (the initialization stage) of ITSHA is to generate an initial solution by the fuzzy c-means (FCM) clustering algorithm \cite{Dunn1973,Bezdek1984} and a random greedy heuristic. %The global structure of $m$TSP instance and the best solution obtained during the procedure of ITSHA are utilized to improve the quality of the initial solution. 
The FCM algorithm and the random greedy heuristic help the algorithm escape from the local optima by providing diverse initial solutions. In the second stage (the improvement stage), a VNS approach based on our proposed neighborhoods is employed to improve the initial solution produced in the first stage. We define a candidate set for each city that records several other nearest cities in ascending order of the distance to reduce the search scope and improve the efficiency of the proposed neighborhoods. The solution can be adjusted several times by exchanging the positions of several cities during the improvement stage to escape from the local optima and find better solutions. ITSHA repeats these two stages until a stopping condition is met.

There are some related studies that apply clustering algorithms to solve the $m$TSP \cite{Latah2016,Lu2016,Xu2018}. These studies all combine clustering algorithms with the population-based algorithms including ACO \cite{Latah2016} and GA \cite{Lu2016,Xu2018} to solve the $m$TSP. Specifically, they apply clustering algorithms to divide the cities into $m$ groups, then use the population-based algorithms to find the shortest $m$ tours that each tour consists of the cities in each group. Obviously, the quality of the results of these algorithms depends too much on the clustering results, since the cities of each tour are fixed according to the clustering result. Their abandoning of the intra-tour improvements results in poor performance. These algorithms also did not compare with other state-of-the-art heuristics on widely used $m$TSP benchmark instances. 

%Thus their performance is not good. 

In our proposed ITSHA algorithm, the clustering algorithm is applied to generate high-quality and diverse initial solutions, which will be improved by the VNS approach in the improvement stage. Both inter-tour and intra-tour improvements are considered by our method. Thus our method can make up for the shortcomings of the related studies that apply clustering algorithms to solve the $m$TSP \cite{Latah2016,Lu2016,Xu2018} described above. Moreover, we tested our algorithm on public \emph{minsum} and \emph{minmax} $m$TSP benchmarks with up to more than 1000 cities. The results show that our ITSHA algorithm significantly outperforms the state-of-the-art heuristics in solving the $m$TSP on both the objectives.

%help initialize the population of the genetic algorithm. However, 

The main contributions of this work are as follows:

\begin{itemize}
\item We propose an iterated two-stage heuristic algorithm called ITSHA to solve the \emph{minsum} and \emph{minmax} $m$TSP with a single depot. ITSHA significantly outperforms the state-of-the-art $m$TSP heuristics, yields 32 new records among 38 public \emph{minsum} $m$TSP instances and 22 new records among 44 public \emph{minmax} $m$TSP instances.
\item We propose three effective and efficient local search operators, called 2-\emph{opt}, \emph{Insert}, and \emph{Swap}, based on the candidate sets. The proposed operators are significantly better than the local search neighborhoods used in existing $m$TSP heuristics.
\item We propose applying the fuzzy clustering algorithm, adjusting the solution and the candidate edges to help the algorithm escape from the local optima and find better results.
\item \textcolor{black}{The proposed local search neighborhoods and the strategies for escaping from the local optima could be applied to other combinatorial optimization problems, such as various variants of the TSP and VRP.}
\end{itemize}

The rest of this paper is organized as follows. Section \ref{sec_ProbDef} formulates the \emph{minsum} and \emph{minmax} $m$TSP. Section \ref{sec_Alg} describes our proposed ITSHA algorithm. Section \ref{sec_Exp} presents experimental results and analyses. Section \ref{sec_Con} contains the concluding remarks.

\section{Problem definition}
\label{sec_ProbDef}
Given a complete undirected graph $G(V,E)$, where $V = \{1,...,n\}$ denotes the set of the cities (note that city 1 is the depot), and $E$ is the pairwise edges $\{e_{ij}| i,j \in V\}$. $c_{ij}$ represents the cost of edge $e_{ij}$ (usually equals to the distance of traveling from city $i$ to city $j$). \textcolor{black}{The $m$TSP is to determine a set of $m$ routes that cover each city exactly once and minimize the objective function (\emph{minsum} or \emph{minmax})}.

Let $x_{ijk}$ be a three-index variable that $x_{ijk} = 1$ when salesman $k$ visits city $j$ immediately after city $i$, otherwise $x_{ijk} = 0$, and $u_i$ be a variable that indicates the visiting rank of city $i$ in order ($u_1 = 0$). The flow based formulation \cite{Christofides1981,Bektas2006} of the \emph{minsum} and \emph{minmax} $m$TSP with the Miller–Tucker–Zemlin (MTZ) \cite{Miller1960} sub-tour elimination constraints is given as follows:

\vspace{1em}
\emph{minsum} $m$TSP:
\begin{equation}
\label{eq_minsum}
\mathrm{Minimize}~~ \sum\limits_{i=1}^n{\sum\limits_{j=1}^n{c_{ij}\sum\limits_{k=1}^m{x_{ijk}}}}
\end{equation}

\emph{minmax} $m$TSP:
\begin{equation}
\label{eq_minmax}
\begin{aligned}
\mathrm{Minimize}~~ &\sum\limits_{i=1}^n{\sum\limits_{j=1}^n{c_{ij}{x_{ijk_{max}}}}},\\
&k_{max}=\mathop{\arg\max}_{k\in\{1,...,m\}}\sum\limits_{i=1}^n{\sum\limits_{j=1}^n{c_{ij}{x_{ijk}}}}
\end{aligned}
\end{equation}

Subject to:
\begin{equation}
\label{eq_c1}
\sum\limits_{i=1}^n{\sum\limits_{k=1}^m{x_{ijk}}}=1,~j=1,...,n
\end{equation}
\vspace{-1em}
\begin{equation}
\label{eq_c2}
\sum\limits_{i=1}^n{x_{ilk}}-\sum\limits_{j=1}^n{x_{ljk}}=0,~k=1,...,m,~l=1,...,n
\end{equation}
\vspace{-1em}
\begin{equation}
\label{eq_c3}
\sum\limits_{j=1}^n{x_{1jk}}=1,~k=1,...,m
\end{equation}
\vspace{-1em}
\begin{equation}
\label{eq_c4}
u_i-u_j+p\sum\limits_{k=1}^m{x_{ijk}}\leq{p-1},~i\neq{j=2,...,n}
\end{equation}
\vspace{-1em}
\begin{equation}
\label{eq_c5}
x_{ijk}\in\{0,1\},~\forall{i,j,k}.
\end{equation}

As shown in Eqs. \ref{eq_minsum} and \ref{eq_minmax}, the \emph{minsum} and \emph{minmax} $m$TSP aim at minimizing the total length of the $m$ tours and the length of the longest tour among all the $m$ tours, respectively. Constraints \ref{eq_c1} state that each city should be visited exactly once and \ref{eq_c2} are the flow conservation constraints that ensure that once a salesman visits a city, then he must also depart from the same city. Constraints \ref{eq_c3} ensure that exactly $m$ salesmen depart from the depot. Constraints \ref{eq_c4} are the extensions of the MTZ \cite{Miller1960} sub-tour elimination constraints to a three-index model, where $p$ denotes the maximum number of cities that any salesman can visit.

\section{The proposed algorithm}
\label{sec_Alg}
The proposed iterated two-stage heuristic algorithm (ITSHA) consists of the initialization stage and the improvement stage. We apply the fuzzy c-means (FCM) clustering algorithm \cite{Dunn1973,Bezdek1984} and a random greedy heuristic to generate an initial solution in the initialization stage. The initialization stage actually helps the algorithm escape from the local optima by providing high-quality and diverse initial solutions. The VNS method based on our proposed neighborhoods is employed in the improvement stage to improve the initial solution. The solution can be adjusted several times by randomly exchanging the positions of several cities to find better solutions. The ITSHA algorithm repeats these two stages until a cut-off time is reached.

This section first describes the main process of the proposed ITSHA algorithm, then introduces the details of the two stages in ITSHA, respectively.

\subsection{Main process of ITSHA}
The main flow of ITSHA is presented in Algorithm \ref{alg_ITSHA}. ITSHA first initializes the candidate set of each city (line 1). We denote $CS$ as the collection of the candidate sets, $CS_i$ as the candidate set of city $i$, and $C_{max}$ (10 by default) as the number of cities in each candidate set, i.e., the initial candidate set of each city records $C_{max}$ other nearest cities in ascending order of the city distance. The search scope of the proposed neighborhoods is significantly reduced by the candidate sets. Thus the efficiency of the algorithm is significantly improved. The approach of applying candidates to reduce the search scope is widely used in the famous TSP heuristics \cite{Helsgaun2000,Nagata2013}. 

\begin{algorithm}[h]
\renewcommand{\algorithmicrequire}{\textbf{Input:}}
\renewcommand{\algorithmicensure}{\textbf{Output:}}
\caption{The ITSHA algorithm}
\label{alg_ITSHA}
\begin{algorithmic}[1]
\REQUIRE the maximum number of the candidate cities: $C_{max}$, the number of times to adjust the solution: $A_t$, the number of the adjusted cities: $A_c$, the cut-off time $t_{max}$, the maximum number of cities that can be visited by any salesman: $p$, the weighting exponent in FCM: $w$, the termination criterion in FCM: $\epsilon$
\ENSURE a solution: $S_{best}$
\STATE $CS:=$ Initialize\_Candidates($C_{max}$)
\STATE Initialize $L(S_{best}):=+\infty$, $L(S)$ is the objective value of solution $S$ (Eq. \ref{eq_minsum} or Eq. \ref{eq_minmax})
\WHILE {the cut-off time $t_{max}$ is not reached}
\STATE Initialize $L(S_{better}):=+\infty$\\
\COMMENT {\emph{the initialization stage}}
\STATE $NC:=$ FCM($w$, $\epsilon$, $p$), $NC_i$ is the group (cluster) that city $i$ belongs to
\STATE $S:=$ Random\_Greedy($NC$, $CS$, $C_{max}$, $S_{best}$)\\
\COMMENT {\emph{the improvement stage}}
\STATE Initialize $num_t:=0$
\WHILE{$num_t < A_t + 1$}
\STATE $S :=$ Adjust\_Solution($S$, $A_c$, $p$) if $num_t > 0$
\STATE $S :=$ VNS($S$, $CS$, $C_{max}$, $p$)
\STATE $S_{better} := S$ if $L(S)<L(S_{better})$
\STATE $num_t:=num_t+1$
\ENDWHILE
\IF{$L(S_{best)}\neq{+\infty}$}
\STATE $CS:=$ Adjust\_Candidates($CS$, $S_{better}$, $S_{best}$)
\ENDIF
\STATE $S_{best} := S_{better}$ if $L(better)<L(S_{best})$
\ENDWHILE
\end{algorithmic}
\end{algorithm}

As shown in Algorithm \ref{alg_ITSHA}, ITSHA repeats the initialization stage (lines 5-6) and the improvement stage (lines 7-13) until the cut-off time $t_{max}$ is reached. The initialization stage generates an initial solution by the FCM algorithm (line 5) and the random greedy function (line 6). The improvement stage applies the VNS method (line 10) based on our proposed neighborhoods to improve the initial solution. The solution obtained during the improvement stage can be adjusted for $A_t$ (3 by default) times to escape from the local optima (line 9). The approach of the solution adjustment function is to randomly delete $A_c$ (5 by default) cities in $S$, then randomly insert these cities into $S$. \textcolor{black}{Note that the adjusted solution should satisfy the constraint that each salesman can visit at most $p$ cities, which is guaranteed by forbidding inserting cities to a tour with $p$ cities.}

Moreover, at the end of each iteration of ITSHA (except the first iteration), the candidate set of each city will be adjusted (lines 14-16). Specifically, if edge $(i,j)$ appears in both $S_{better}$ and $S_{best}$, the last candidate city in $CS_i$ will be replaced with city $j$ if $j\notin{CS_i}$. The method of adjusting candidate cities allows the search neighborhoods to change adaptively to enhance the algorithm's robustness and search ability. The experimental results also demonstrate that adjusting candidate cities can improve the performance of our ITSHA algorithm.

\subsection{The initialization stage of ITSHA}
This subsection introduces the initialization stage of ITSHA. We describe the process of the FCM algorithm and the random greedy function, respectively.

\subsubsection{Fuzzy c-means clustering}
The fuzzy c-means clustering (FCM) algorithm \cite{Dunn1973,Bezdek1984} is an important vehicle to cope with overlapping clustering, which uses a membership matrix $U=[u_{ij}] \in \mathbb{R}^{\mathrm{N} \times \mathrm{c}}$ to represent the result of clustering $N$ elements into $c$ clusters. The membership degree $u_{ij}$ of the element $i$ in the cluster $j$ is subjected to the following constraints: 1) $u_{ij} \in [0,1]$. 2) $\sum\nolimits_{j=1}^c{u_{ij}} = 1$. 

In our ITSHA algorithm, we apply FCM to cluster the $n-1$ cities (all the $n$ cities except the depot) into $m$ clusters according to their positions, so as to assign the cities of each cluster to a salesman. With the help of the clustering algorithm, the positions of the cities assigned to each salesman (in each cluster) are close. We employ the FCM algorithm rather than the common c-means algorithm used in \cite{Latah2016,Lu2016,Xu2018} since the randomness and robustness of FCM are better than those of c-means. In other words, the diversity of the initial solutions generated by FCM, are better than by c-means. The goal of the FCM algorithm in ITSHA is to minimize the following objective function $J$:
%\vspace{-0.5em}
\begin{equation}
\label{eq_J}
J = \sum_{i=2}^{n}{\sum_{j=1}^{m}{u_{ij}^w{\left\|x_i-v_j\right\|^2}}},
\end{equation}
where $w$ is the weighting exponent, $x_i$ is the position of city $i$, $v_j$ is the cluster center of the cluster $j$. The FCM is carried out through iterative minimization of the objective function $J$ through updating the cluster center $v_j$ and the membership degrees $u_{ij}$ according to the following formulas:
%\vspace{-0.5em}
\begin{equation}
\label{eq_V}
v_j = \frac{\sum\nolimits_{i=2}^{n}{u_{ij}^w{x_i}}}{\sum\nolimits_{i=2}^{n}{u_{ij}^w}},~j=1,...,m
\end{equation}
and
\begin{equation}
\label{eq_U}
\begin{aligned}
u_{ij} = &\frac{1}{\sum\nolimits_{k=1}^m{(\frac{\left\|x_i-v_j\right\|}{\left\|x_i-v_k\right\|})^{\frac{2}{w-1}}}},\\
&i=2,...,n,~j=1,...,m.
\end{aligned}
\end{equation}

The flow of the fuzzy clustering procedure in our ITSHA algorithm is shown in Algorithm \ref{alg_FCM}. The FCM algorithm first randomly initializes the membership matrix $U$ (line 3). The random initialized membership matrix can help the initialization stage of ITSHA generate diverse initial solutions. FCM then repeatedly updates the clustering centers and membership matrix until the membership matrix converges (lines 4-7). Lines 9-12 can prevent the clustering algorithm from obtaining empty clusters. In general, city $i$ will be assigned to the salesman (cluster) $t$ that the membership degree $u_{it}$ is the largest among $u_{ij}, j\in{\{1,...,m\}}$ (lines 17-18). A cluster with $p-1$ cities can not be assigned more cities since each salesman can visit at most $p$ cities include the depot (line 17).

\begin{algorithm}[h]
\renewcommand{\algorithmicrequire}{\textbf{Input:}}
\renewcommand{\algorithmicensure}{\textbf{Output:}}
\caption{Fuzzy clustering procedure}
\label{alg_FCM}
\begin{algorithmic}[1]
\REQUIRE the weighting exponent in FCM: $w$, the termination criterion in FCM: $\epsilon$, the maximum number of cities that can be visited by any salesman: $p$
\ENSURE a vector that represents the cluster each city belongs to: $NC$
\STATE Initialize $NC_i=0$ for each $i\in{\{2,...,n\}}$
\STATE Let $C_{num}:=[C_{{num}_j}], j\in{\{1,...,m\}}$ be a vector that $C_{{num}_j}$ represents the number of the cities belong to cluster $j$
\STATE Let $k:=1$, and randomly initialize the membership matrix $U^k$
\REPEAT
\STATE Compute $v_j, j\in{\{1,...,m\}}$ according to Eq. \ref{eq_V}
\STATE Compute the updated membership matrix $U^{k+1}$ according to Eq. \ref{eq_U}, $k:=k+1$
\UNTIL{$\max_{i,j}{\{|u_{ij}^{k}-u_{ij}^{k-1}|\}}<\epsilon$}
\STATE $U:=U^k$
\FOR{$j:=1$ to $m$}
\STATE $t:=\arg\max_{i\in{\{2,...,n\}} \wedge {NC_i = 0}}{u_{ij}}$
\STATE $NC_t:=j$, $C_{{num}_j}:=1$
\ENDFOR
\FOR{$i:=2$ to $n$}
\IF{$NC_i\neq{0}$}
\STATE \textbf{continue}
\ENDIF
\STATE $t:=\arg\max_{j\in{\{1,...,m\}} \wedge C_{{num}_j}<p-1}{u_{ij}}$
\STATE $NC_i:=t$, $C_{{num}_t}:=C_{{num}_t}+1$
\ENDFOR
\end{algorithmic}
\end{algorithm}

%In summary, we apply the FCM algorithm in ITSHA to divide the cities into $m$ clusters according to their positions, so as to assign the cities of each cluster to each salesman. Intuitively, with the help of the clustering algorithm, the positions of the cities assigned to each salesman are close. Thus the initial We employ the FCM algorithm rather than the common c-means algorithm used in \cite{Latah2016,Lu2016,Xu2018} since the randomness and robustness of FCM are better than those of c-means.

\subsubsection{Random greedy function}
We propose a random greedy function to generate a feasible initial solution for the $m$TSP based on the clustering results of FCM. The process of the random greedy function is shown in Algorithm \ref{alg_Greedy}. The random greedy function actually determines a connected sequence of the cities in each cluster coupled with the depot (i.e., $C^j$ in line 1) to produce a solution. For generating each of the $m$ tours, a random city $sc$ is chosen firstly (line 4), and the current city $cc$ is set to be $sc$. Then, as long as not all cities in $C^j$ have been chosen, choose the next city $nc$ to follow $cc$ in the current tour, and set $cc$ equal to $nc$. $nc$ is chosen as follows: 

(1) If possible, choose $nc$ such that $(nc,cc)$ is an edge of the best solution $S_{best}$ (lines 8-13). 

(2) Otherwise, if possible, choose $nc$ from the candidate set of $cc$, $CS_{cc}$ (lines 14-18). 

(3) Otherwise, choose $nc$ at random among those cities not already chosen in $C^j$ (lines 19-23).

When more than one city may be chosen, the city is chosen at random among the alternatives (a one-way list of cities, as shown in line 24 in Algorithm \ref{alg_Greedy}). The $m$ sequences of chosen cities constitute the initial solution $S$ of $m$TSP.

\begin{algorithm}[h]
\renewcommand{\algorithmicrequire}{\textbf{Input:}}
\renewcommand{\algorithmicensure}{\textbf{Output:}}
\caption{Random greedy procedure}
\label{alg_Greedy}
\begin{algorithmic}[1]
\REQUIRE the clustering results: $NC$, the candidate sets: $CS$, the maximum number of the candidate cities: $C_{max}$, the best solution: $S_{best}$
\ENSURE a solution: $S$
\STATE Let $C^j=\{1\}\cup{\{i|i\in{\{2,...,n\}} \wedge NC_i=j}\}$ be a set of the cities assigned to salesman $j$ (i.e., the cities in cluster $j$ coupled with the depot), $j\in{\{1,...,m\}}$
%that $Se_i=0$ indicates city $i$ has not been visited in the initial solution, otherwise $Se_i=1$,
\STATE Initialize $Se \in \{0\}^{n}$, $Se_i=1$ indicates city $i$ has been selected, otherwise $Se_i=0$ 
\FOR{$j:=1$ to $m$}
\STATE $Se_1:=0$
\STATE Randomly select a starting city $sc$ in $C^j$, $Se_{sc}:=1$, $C^j:=C^j\backslash\{sc\}$, set current city $cc:=sc$ 
\REPEAT
\STATE Initialize a set of the alternative cities $AC:=\emptyset$
\IF{$L(S_{best})\neq{+\infty}$}
\STATE Let $a_1,a_2$ be the two cities connected with $cc$ in $S_{best}$. 
\FOR{$i:=1$ to 2}
\STATE If $NC_{a_i}=j \wedge Se_{a_i}=0$, $AC:=AC\cup{\{{a_i}\}}$
\ENDFOR
\ENDIF
\IF{$|AC|=0$}
\FOR{city $k\in{CS_{cc}}$}
\STATE If $NC_k=j \wedge Se_k=0$, $AC:=AC\cup{\{{k}\}}$
\ENDFOR
\ENDIF
\IF{$|AC|=0$}
\FOR{city $k\in{C^j}$}
\STATE If $NC_k=j \wedge Se_k^j=0$, $AC:=AC\cup{\{{k}\}}$
\ENDFOR
\ENDIF
\STATE Randomly select next city $nc\in{\{AC\}}$, $Se_{nc}:=1$, $C^j:=C^j\backslash\{nc\}$, connect city $cc$ and $nc$ in the solution $S$, $cc:=nc$
\UNTIL{$C^j=\emptyset$}
\STATE Connect city $cc$ and $sc$ in the solution $S$
\ENDFOR
\end{algorithmic}
\end{algorithm}

In summary, the random greedy function constructs the initial solution iteratively. In each iteration, the procedure tries to let the salesman follow the current best tour or select the next cities from the candidate sets. That is, the global structure of the $m$TSP instance (i.e., the candidate sets) and the best solution obtained during the procedure of ITSHA are utilized to improve the quality of the initial solution. The randomness of FCM and the random greedy function guarantees the diversity of the initial solutions. In a word, the initialization stage of ITSHA can provide high-quality and diverse initial solutions to help the algorithm escape from the local optima and yield better results.

\subsection{The improvement stage of ITSHA}
This subsection mainly describes the process of the variable neighborhood search (VNS) in the improvement stage of ITSHA. We first introduce the neighborhoods used in our algorithms, then present the flow of the VNS.

\subsubsection{Neighborhoods used in ITSHA}
\begin{figure*}[t]
\centering
\subfigure[2-\emph{opt} move]{
\includegraphics[width=0.6\columnwidth]{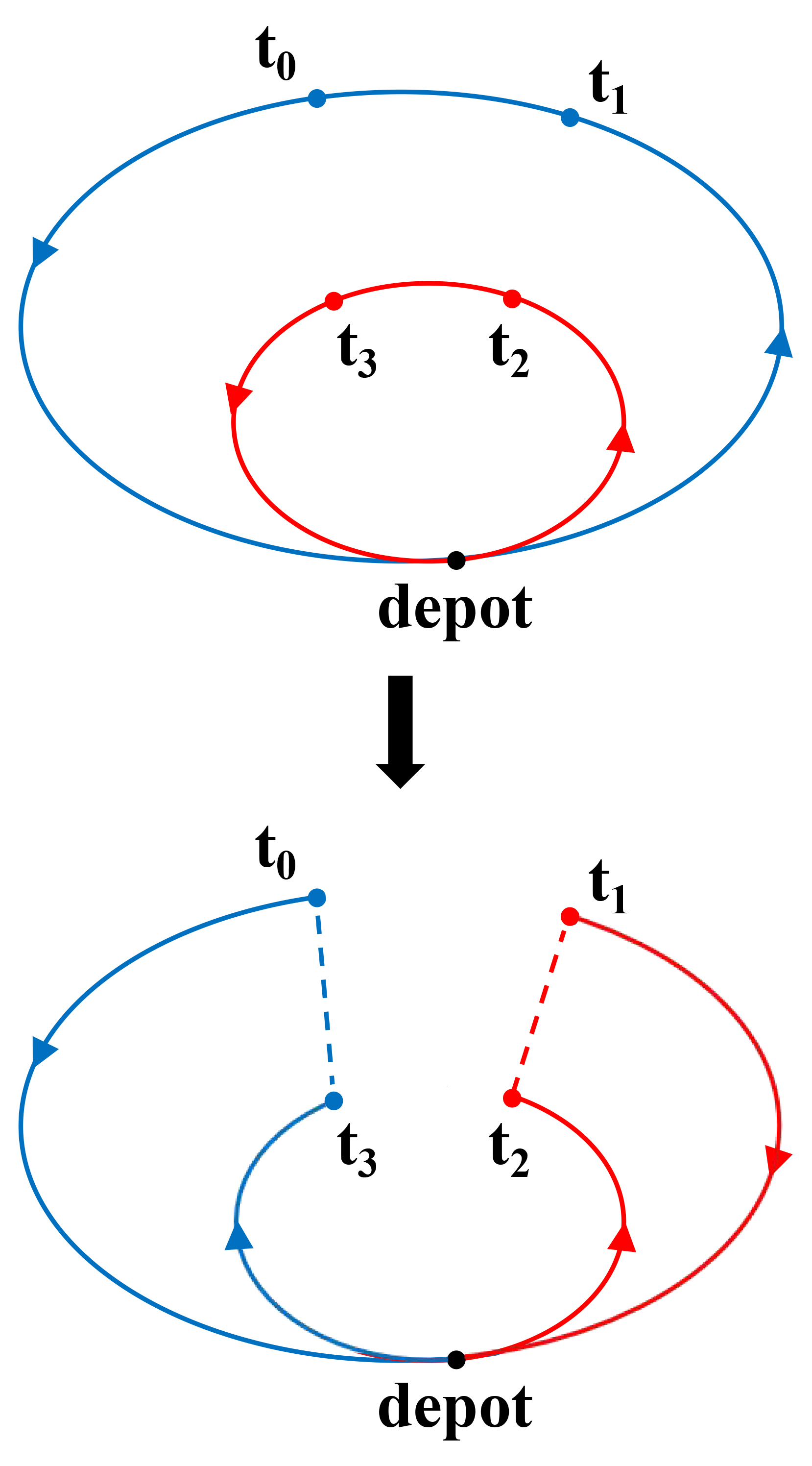}  
\label{fig_2opt}}%\hspace{2mm}
\hspace{0.5em}
\subfigure[\emph{Insert} move]{
\includegraphics[width=0.6\columnwidth]{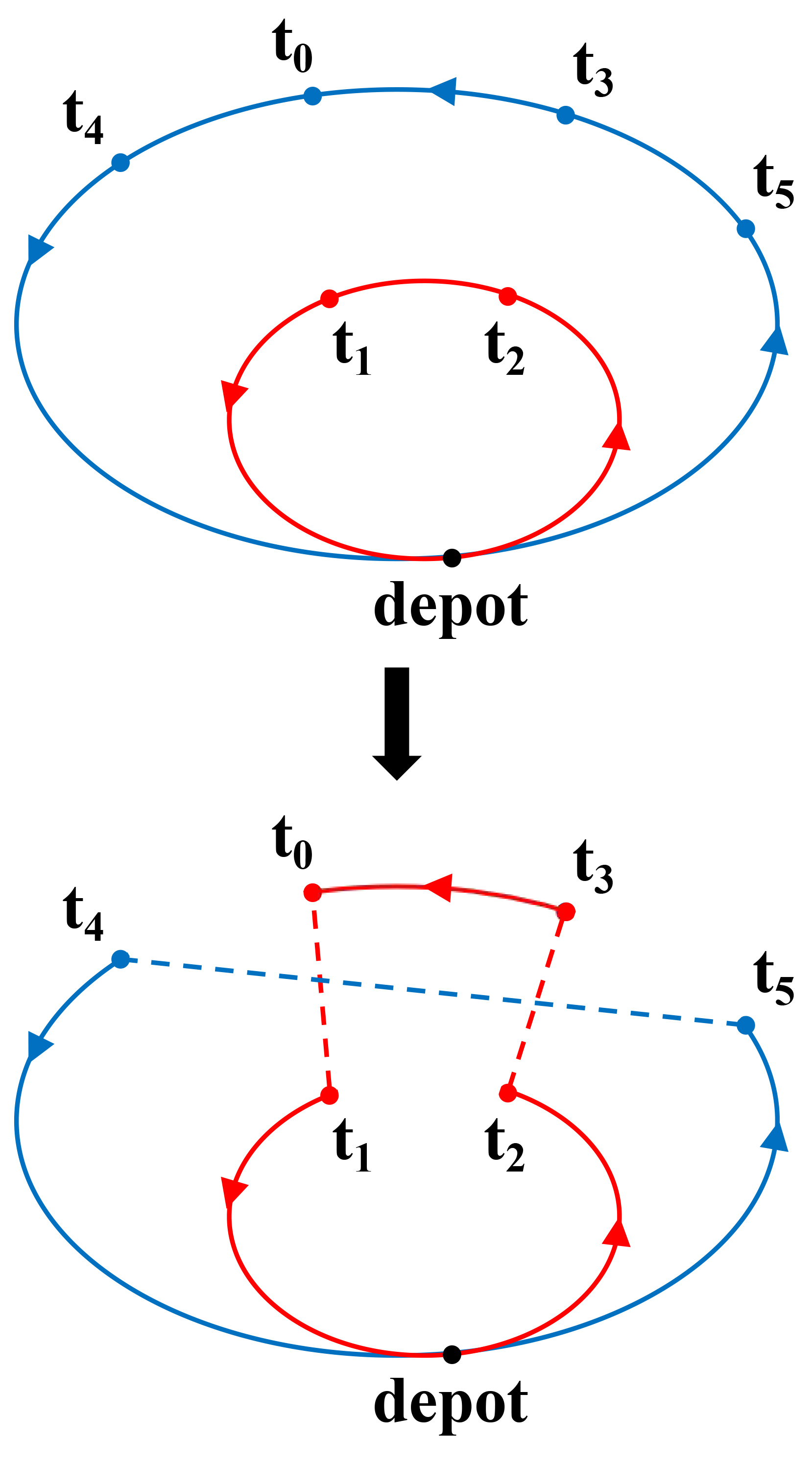} 
\label{fig_Insert}}
\hspace{0.12em}
\subfigure[\emph{Swap} move]{
\includegraphics[width=0.6\columnwidth]{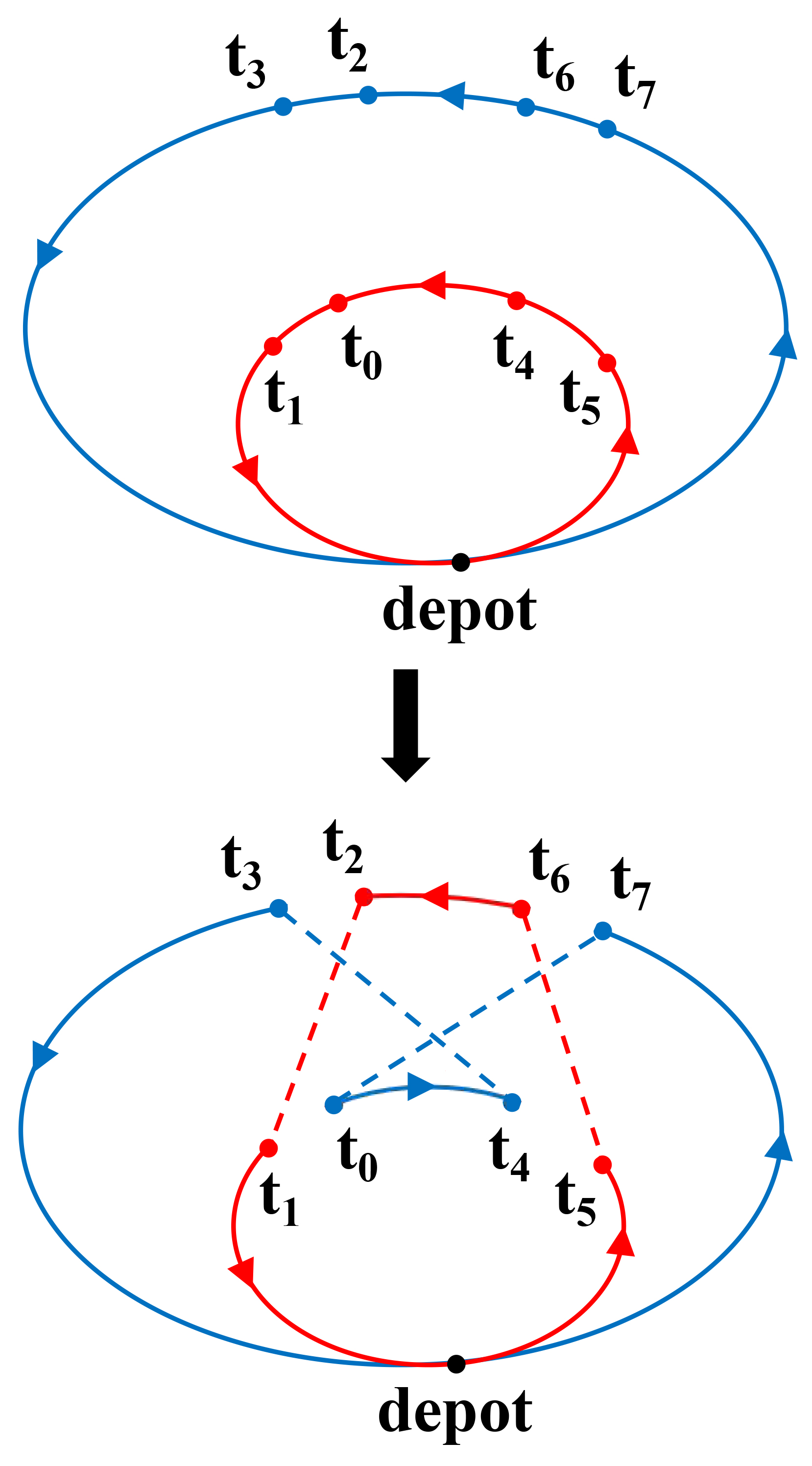} 
\label{fig_Swap}}
\caption{An Illustration of the neighborhoods used in the VNS process of ITSHA. Note that the routes of different salesmen are in different colors. An arc with an arrow is a part of a tour, that may contain multiple cities. An arc without an arrow is an edge of the tour.}
\label{fig_VNS}
\end{figure*}

The VNS approach \cite{Mladenovic1997} is widely used in the routing problems include TSP \cite{Hore2018}, VRP \cite{Kytojoki2007,Defryn2017}, and $m$TSP \cite{Soylu2015,Wang2017}. The performance of VNS strongly depends on the design of the neighborhood structures. However, the existing neighborhoods used in the state-of-the-art $m$TSP heuristics \cite{Soylu2015,Wang2017,Karabulut2021} are inefficient, \textcolor{black}{since they contain lots of low-quality operators that should not be considered.} To handle this problem, we introduce three neighborhoods, 2-\emph{opt}, \emph{Insert} and \emph{Swap}, applied in our ITSHA algorithm that are significantly more effective and efficient than the neighborhoods used in \cite{Soylu2015,Karabulut2021,Wang2017}. The three neighborhoods are illustrated in Figure \ref{fig_VNS}.

In Figure \ref{fig_VNS}, the routes of different salesmen are in different colors. It is worth mentioning that, an arc with an arrow is a part of a tour, that may contain multiple cities. An arc without an arrow is an edge of the tour. The detailed description of the three neighborhoods is as follows. 

(1) 2-\emph{opt}:
The 2-\emph{opt} operator is shown in Figure \ref{fig_2opt}. It tries to replace two edges in the current $m$TSP tour, $(t_0,t_1)$ and $(t_2,t_3)$, with two new edges, $(t_0,t_3)$ and $(t_1,t_2)$, to improve the solution. However, performing an exhaustive search of 2-\emph{opt} is too time consuming. Therefore, we restrict that $t_2$ must be selected in the candidate set of $t_1$, which can reduce the search scope of $t_2$ from about $n$ to $C_{max}$.

(2) \emph{Insert}:
The \emph{Insert} operator is shown in Figure \ref{fig_Insert}. It tries to insert a sequence of cities between $t_0$ and $t_3$ into an edge $(t_1,t_2)$ to improve the solution. We restrict that $t_1$ must be selected in the candidate set of $t_0$, and $t_3$ must be selected in the candidate set of $t_2$. \textcolor{black}{Such an restriction can reduce the search scope from about $n^2$ to $C_{max}^2$.} It is worth mentioning that, the operators \emph{One-point} move, \emph{Or-opt}2 move, \emph{Or-opt}3 move in \cite{Soylu2015,Wang2017}, and the \emph{Or-opt}4 move in \cite{Wang2017} are the special cases of our \emph{Insert} operator when $C_{max}$ is set to $n$ and the sequence between $t_0$ and $t_3$ contains one, two, three, four cities respectively. %And the $N_1$ and $N_2$ operators in \cite{Karabulut2021} are the special case of our \emph{Insert} operator when the sequence between $t_0$ and $t_3$ contains one city.

(3) \emph{Swap}:
The \emph{Swap} operator is shown in Figure \ref{fig_Swap}. It tries to improve the solution by swapping two sequences of cities, one of them contains the cities between $t_0$ and $t_4$, another contains the cities between $t_2$ and $t_6$. We restrict that $t_2$ must be selected in the candidate set of $t_1$, $t_4$ must be selected in the candidate set of $t_3$, and $t_6$ must be selected in the candidate set of $t_5$. \textcolor{black}{Such an restriction can reduce the search scope from about $n^3$ to $C_{max}^3$.} It is worth mentioning that, the operators \emph{Two-point} move and \emph{Three-point} move in \cite{Soylu2015,Wang2017} are the special cases of our \emph{Swap} operator when $C_{max}$ is set to $n$, $t_0$ is equal to $t_4$, and the sequence between $t_2$ and $t_6$ contains one, two cities respectively. %And the $N_3$ and $N_4$ operators in \cite{Karabulut2021} are the special case of our \emph{Swap} operator when the sequence between $t_2$ and $t_6$ contains one city.

Moreover, each of the three operators used in ITSHA can be applied for an inter-tour or an intra-tour improvement. Such a mechanism is more effective than the VNS approaches in the state-of-the-art heuristics \cite{Soylu2015,Wang2017} that only apply the 2-\emph{opt} operator to perform inter-tour improvements, and other operators to perform intra-tour improvements. 

\subsubsection{The process of VNS}
The process of the VNS local search in ITSHA is shown in Algorithm \ref{alg_VNS}. The VNS sequentially applies the three operators to improve the input solution to a local optimum. Specifically, the function \emph{Insert}() in line 3 improves the input solution $S_{old}$ to a local optimum for the \emph{Insert} operator, i.e., \emph{Insert}() tries to improve the solution until no \emph{Insert} operator can be found to improve the current solution. The \emph{Swap} operator and the 2-\emph{opt} operator are applied in the same way (lines 4-5). Note that for the \emph{minsum} $m$TSP, a solution is considered to be improved if the objective value (Eq. \ref{eq_minsum}) is reduced. For the \emph{minmax} $m$TSP, a solution is considered to be improved if the objective value (Eq. \ref{eq_minmax}) is reduced, or the objective value is unchanged and the total length of the $m$ tours is reduced. The solutions obtained during the VNS process are all feasible, i.e., they all satisfy the constraint that each salesman can visit at most $p$ cities. \textcolor{black}{The feasibility of the solutions is guaranteed by abandoning the operators that result in infeasible solutions.} The VNS process terminates when the current solution can not be improved by any of the three operators (line 6).

\begin{algorithm}[h]
\renewcommand{\algorithmicrequire}{\textbf{Input:}}
\renewcommand{\algorithmicensure}{\textbf{Output:}}
\caption{VNS in ITSHA}
\label{alg_VNS}
\begin{algorithmic}[1]
\REQUIRE a solution: $S$, the candidate sets: $CS$, the maximum number of the candidate cities: $C_{max}$, the maximum number of cities that can be visited by any salesman: $p$
\ENSURE a solution: $S$
\REPEAT
\STATE $S_{old}:=S$
\STATE $S_{old}:=$ \emph{Insert}($S_{old}$, $CS$, $C_{max}$, $p$)
\STATE $S_{old}:=$ \emph{Swap}($S_{old}$, $CS$, $C_{max}$, $p$)
\STATE $S:=$ 2-\emph{opt}($S_{old}$, $CS$, $C_{max}$, $p$)
\UNTIL{$S={S_{old}}$}
\end{algorithmic}
\end{algorithm}

%The three operators try to improve current solution by performing a corresponding move. For the \emph{minsum} $m$TSP, a solution is considered to be improved if the objective value (Eq. \ref{eq_minsum}) is reduced. And for the \emph{minmax} $m$TSP, a solution is considered to be improved if the objective value (Eq. \ref{eq_minmax}) is reduced or the objective value is unchanged and the total length of the $m$ tours is reduced. If a move can improve current solution and satisfies the constraint that each salesman can visit at most $p$ cities, then the algorithm changes current solution by performing this move. Each of the above three operators (2-\emph{opt}, \emph{Insert}, \emph{Swap}) tries to improve the solution to the local optimal. 

With the help of the restriction of the search scope based on the candidate sets, \textcolor{black}{the operators are refined by abandoning lots of low-quality moves.} As a result, the computational complexity of improving a solution to the local optimum for each of the three operators (2-\emph{opt}, \emph{Insert}, \emph{Swap}) is $O(n)$ (i.e., the computational complexity of line 3/4/5 in Algorithm \ref{alg_VNS} is $O(n)$). This is much more efficient than the neighborhoods used in \cite{Soylu2015,Wang2017}, where the computational complexity of applying an operator once is $O(n^2)$.

In summary, the VNS process in our ITSHA algorithm is significantly better than the VNS used in \cite{Soylu2015,Wang2017} because: \textcolor{black}{1) Our neighborhoods are much more efficient than theirs, since the low-quality moves can be refined by applying the candidate sets. 2) Our neighborhoods are much more effective than theirs, since our operators \emph{Insert} and \emph{Swap} can move a sequence of cities, which leads to a wide and deep search region. Thus our method can find higher-quality solutions.} 3) All of our neighborhoods can be used to perform inter-tour or intra-tour improvements, while they only apply the 2-\emph{opt} operator to perform inter-tour improvements, and other operators to perform intra-tour improvements.

%\subsubsection{Solution adjustment}
%In order to escape from the local optimal, we propose to use a solution adjustment function to adjust the output solution $S$ of the VNS process. The approach of the solution adjustment function is to randomly delete $A_c$ cities in $S$, then randomly insert these cities into the solution. Note that the adjusted solution should satisfies the constraint that each salesman can visit at most $p$ cities.

\section{Experimental results}
\label{sec_Exp}
Experimental results provide insight on why and how the proposed approach ITSHA is effective, suggesting that the VNS based on the proposed neighborhoods is efficient and effective. The fuzzy clustering algorithm, the solution adjustment process and the approach of adjusting candidate sets can help the algorithm escape from the local optima and find better solutions. 

In this section, we first present the the benchmark instances, baseline algorithms, experimental setup and various variants of ITSHA, then present and analyze the experimental results.

\subsection{Benchmark instances}
\label{sec_instances}
We test our ITSHA algorithm in solving the \emph{minsum} and \emph{minmax} $m$TSP on the benchmark instances used in the state-of-the-art heuristics \cite{Venkatesh2015,Soylu2015,Lu2019,Karabulut2021,Lo2018,Wang2017}, a total of 38 for the \emph{minsum} $m$TSP with the number of cities ranges from 11 to 1002, and 44 for the \emph{minmax} $m$TSP with the number of cities ranges from 11 to 1173. Note that for all the tested instances, the number in an instance name indicates the number of cities in that instance, and the first city of an instance is set to be the depot. We divide these instances into the following four sets:

\begin{itemize}
\item \textit{Set \uppercase\expandafter{\romannumeral1}}: This set contains a total of 8 instances. Among them, there are three instances with $n$ = 128 and $m$ = 10, 15, 30 (denoted as \emph{128}), and five small instances called \emph{11a}, \emph{11b}, \emph{12a}, \emph{12b}, and \emph{16} with $m$ = 3. Instances \emph{11a}, \emph{12a} and \emph{16} comprise the first 11, 12, and 16 cities of the $n$ = 51 instance of \cite{Carter2006} respectively, whereas \emph{11b}, \emph{12b}, and \emph{128} instances are derived from sp11, uk12, and sgb128 data sets\footnote{https://people.sc.fsu.edu/\%7Ejburkardt/datasets/cities/cities.html}. The 8 instances in this set are used in \cite{Venkatesh2015,Soylu2015,Karabulut2021} for the \emph{minsum} and \emph{minmax} $m$TSP.
\item \textit{Set \uppercase\expandafter{\romannumeral2}}: This set contains a total of 12 instances that consist of three symmetric TSP instances \emph{eil51}, \emph{kroD100} and \emph{mTSP150} in the TSPLIB\footnote{http://comopt.ifi.uni-heidelberg.de/software/TSPLIB95}. Among these 12 instances, there are three instances with $n$ = 51 and $m$ = 3, 5, 10, four instances with $n$ = 100 and $m$ = 3, 5, 10, 20, and five instances with $n$ = 150 and $m$ = 3, 5, 10, 20, 30. This set of instances is widely used in \cite{Venkatesh2015,Soylu2015,Lu2019,Karabulut2021} for the \emph{minsum} and \emph{minmax} $m$TSP.
\item \textit{Set \uppercase\expandafter{\romannumeral3}}: This set contains a total of 18 instances that consist of six symmetric TSP instances \emph{pr76}, \emph{pr152}, \emph{pr226}, \emph{pr299}, \emph{pr439}, and \emph{pr1002} in the TSPLIB. The number of salesmen is set to be $m$ = 5, 10, 15 for these six TSP instances. The maximum number of cities that can be visited by any salesman is set to be $p=20/40/50/70/100/220$ for the instances in this set with $n=76/152/226/299/439/1002$. This set of instances is used in \cite{Lo2018} for the \emph{minsum} $m$TSP.
\item \textit{Set \uppercase\expandafter{\romannumeral4}}: This set contains a total of 24 instances that consist of six symmetric TSP instances \emph{ch150}, \emph{kroA200}, \emph{lin318}, \emph{att532}, \emph{rat783}, and \emph{pcb1173} in the TSPLIB. The number of salesmen is set to be $m$ = 3, 5, 10, 20 for these six TSP instances. This set of instances is used in \cite{Karabulut2021,Wang2017} for the \emph{minmax} $m$TSP.
\end{itemize}

\begin{table*}[t]
\caption{Information of the baseline algorithms and our ITSHA algorithm.}
\label{table_informs}
\scalebox{0.7}{\begin{tabular}{lrrrr} \toprule
Method   & Objectives    & Benchmarks   & Stopping criterion                        & Computer \\ \hline
ABC(FC)  & \emph{minsum} and \emph{minmax} & \textit{Sets} \textit{\uppercase\expandafter{\romannumeral1}} and \textit{\uppercase\expandafter{\romannumeral2}} & The maximum number of iterations is 1000    & 2.83 GHz Core 2 Quad system \\
ABC(VC)  & \emph{minsum} and \emph{minmax} & \textit{Sets} \textit{\uppercase\expandafter{\romannumeral1}} and \textit{\uppercase\expandafter{\romannumeral2}} & The maximum number of iterations is 1000    & 2.83 GHz Core 2 Quad system \\
IWO      & \emph{minsum} and \emph{minmax} & \textit{Sets} \textit{\uppercase\expandafter{\romannumeral1}} and \textit{\uppercase\expandafter{\romannumeral2}} & The maximum number of iterations is 1000    & 2.83 GHz Core 2 Quad system \\
GVNS     & \emph{minsum} and \emph{minmax} & \textit{Sets} \textit{\uppercase\expandafter{\romannumeral1}} and \textit{\uppercase\expandafter{\romannumeral2}} & The cut-off time is $n$ seconds               & 2.4 GHz
workstation  \\
ACO & \emph{minsum} and \emph{minmax} & \textit{Set} \textit{\uppercase\expandafter{\romannumeral2}} & The maximum number of iterations is 10$n$     & --       \\
ES       & \emph{minsum} and \emph{minmax}   & \textit{Sets} \textit{\uppercase\expandafter{\romannumeral1}}, \textit{\uppercase\expandafter{\romannumeral2}} and \textit{\uppercase\expandafter{\romannumeral4}}       & The cut-off time is $n$ seconds for \textit{Sets} \textit{\uppercase\expandafter{\romannumeral1}} and \textit{\uppercase\expandafter{\romannumeral2}}, $n$/5 seconds for \textit{Set} \textit{\uppercase\expandafter{\romannumeral4}}            & Intel Core2 Quad Q9400 CPU with 2.66 GHz  \\
GAL      & \emph{minsum}      & \textit{Set} \textit{\uppercase\expandafter{\romannumeral3}}     & When the population converges              &  Intel Core i7-3370 CPU @3.4 GHz  \\
MASVND   & \emph{minmax}      & \textit{Set} \textit{\uppercase\expandafter{\romannumeral4}}      & The cut-off time is $n$/5 seconds             & Intel Core i7-3770 CPU @ 3.40 GHz
%ITSHA    & \emph{minsum} and \emph{minmax} & The cut-off time is $n$ seconds for Sets \uppercase\expandafter{\romannumeral1} and \uppercase\expandafter{\romannumeral2}, $n$/5 seconds for Set \uppercase\expandafter{\romannumeral4}               & 2.00 GHz 
\\ \bottomrule
\end{tabular}}
\end{table*}

\subsection{Baseline algorithms}
We compare the ITSHA with the state-of-the-art $m$TSP heuristics. For example, the artificial bee colony (ABC) algorithms proposed by Venkatesh and Singh \cite{Venkatesh2015}, denoted as ABC(FC) and ABC(VC), which represent that a parameter in the ABC algorithm is fixed or variable, respectively. Venkatesh and Singh \cite{Venkatesh2015} also propose an invasive weed optimization (IWO) algorithm. Moreover, the general variable neighborhood search (GVNS) heuristic proposed by Soylu \cite{Soylu2015}, the ant colony optimization (ACO) algorithm proposed by Lu and Yue \cite{Lu2019}, the evolution strategy approach called ES proposed by Karabulut et al. \cite{Karabulut2021}. The above six algorithms can be applied to solve both of the \emph{minsum} and the \emph{minmax} $m$TSP. In addition, there are some state-of-the-art heuristics aim at solving one of the \emph{minsum} and the \emph{minmax} $m$TSP. For example, the genetic algorithm for the \emph{minsum} $m$TSP called GAL proposed by Lo et al. \cite{Lo2018}, and the memetic algorithm for the \emph{minmax} $m$TSP called MASVND proposed by Wang et al. \cite{Wang2017}.

In summary, we compare the ITSHA with the above eight heuristics: ABC(FC), ABC(VC), IWO \cite{Venkatesh2015}, GVNS \cite{Soylu2015}, ACO \cite{Lu2019}, ES \cite{Karabulut2021}, GAL \cite{Lo2018}, and MASVND \cite{Wang2017}. Note that the results of these algorithms are all from the literature. 

\textcolor{black}{The information of these eight algorithms includes the objectives of the problems (\emph{minsum} and \emph{minmax}) that they can solve, the benchmarks they used, the stopping criteria, as well as the systems on which the algorithms run is concluded in Table \ref{table_informs}.}

%For baseline algorithms in the comparison, we choose the following state-of-the-art $m$TSP heuristics: (1) the artificial bee colony algorithm with a fixed parameter proposed by Venkatesh and Singh \cite{Venkatesh2015}, denoted as ABC(FC). (2) the artificial bee colony algorithm with a variable parameter proposed by Venkatesh and Singh \cite{Venkatesh2015}, denoted as ABC(VC). (3) the invasive weed optimization (IWO) algorithm \cite{Venkatesh2015}. (4) the general variable neighborhood search (GVNS) heuristic \cite{Soylu2015}. (5)

\subsection{Experimental setup}
Our proposed algorithm ITSHA was coded in the C++ Programming Language and was performed on a server with Intel® Xeon® E5-2640 v2 2.00 GHz 8-core CPU and 64 GB RAM. \textcolor{black}{Note that the machine we used is worse than the machines listed in Table \ref{table_informs}.} The parameters in ITSHA are set as follows: $C_{max}=10$, $A_t=3$, $A_c=5$, $w=2$, $\epsilon = 0.0001$. All these parameter values have been chosen empirically. \textcolor{black}{To have a fair comparison with the baseline algorithms, ITSHA has been run for 10 independent replications for the instances in \textit{Sets} \textit{\uppercase\expandafter{\romannumeral1}} and \textit{\uppercase\expandafter{\romannumeral2}} as \cite{Soylu2015} did, and 20 independent replications for the instances in \textit{Sets} \textit{\uppercase\expandafter{\romannumeral3}} and \textit{\uppercase\expandafter{\romannumeral4}} as \cite{Karabulut2021,Lo2018,Wang2017} did. The cut-off time of the instances in \textit{Sets} \textit{\uppercase\expandafter{\romannumeral1}}, \textit{\uppercase\expandafter{\romannumeral2}}, and \textit{\uppercase\expandafter{\romannumeral3}} is set to $n$ seconds as \cite{Soylu2015,Karabulut2021} did, and $n$/5 seconds as \cite{Karabulut2021,Wang2017} did.}

%each instance in \textit{Set} \uppercase\expandafter{\romannumeral1} is solved 30 times by ITSHA  

%Each $m$TSP instance is solved 10 times by ITSHA in the comparison, each time with a different random seed. %\textcolor{black}{The cut-off time of ITSHA is set to $n$ seconds as GVNS \cite{Soylu2015} and ES \cite{Karabulut2021} do}.

\begin{table*}[t]
\caption{Comparison of ITSHA and the baseline algorithms in solving the \emph{minsum} and \emph{minmax} $m$TSP on the instances of \textit{Set \uppercase\expandafter{\romannumeral1}}. Best results appear in bold.}
\label{table_set1}
\begin{tabular}{lrrrrrrrr} \toprule
\multirow{2}{*}{Instance} & \multirow{2}{*}{$m$} & \multirow{2}{*}{ABC(FC)} & \multirow{2}{*}{ABC(VC)} & \multirow{2}{*}{IWO} & \multirow{2}{*}{GVNS} & \multirow{2}{*}{ES} & \multicolumn{2}{r}{ITSHA} \\ \cline{8-9} 
                          &                    &                          &                          &                      &                       &                     & Best            & Average \\ \hline
\multicolumn{9}{l}{\emph{minsum}}                                                                                                                                                                            \\ \hline
11a                       & 3                  & 198                      & 198                      & 198                  & 198                   & 198                 & \textbf{196}    & 196.0   \\
11b                       & 3                  & \textbf{135}             & \textbf{135}             & \textbf{135}         & \textbf{135}          & \textbf{135}        & \textbf{135}    & 135.0   \\
12a                       & 3                  & 199                      & 199                      & 199                  & 199                   & 199                 & \textbf{198}    & 198.0   \\
12b                       & 3                  & \textbf{2295}            & \textbf{2295}            & \textbf{2295}        & \textbf{2295}         & \textbf{2295}       & \textbf{2295}   & 2295.0  \\
16                        & 3                  & 242                      & 242                      & 242                  & 242                   & 242                 & \textbf{241}    & 241.0\vspace{0.5em}    \\
128                       & 10                 & 30799                    & 26482                    & 24514                & 22647                 & 21354               & \textbf{21113}  & 21160.3 \\
                          & 15                 & 32777                    & 28405                    & 26368                & 25204                 & 23962               & \textbf{23838}  & 23896.6 \\
                          & 30                 & 43599                    & 41754                    & 39579                & 37383                 & 36871               & \textbf{36655}  & 36726.3 \\ \hline
\multicolumn{9}{l}{\emph{minmax}}                                                                                                                                                                            \\ \hline
11a                       & 3                  & \textbf{77}              & \textbf{77}              & \textbf{77}          & \textbf{77}           & \textbf{77}         & \textbf{77}     & 77.0    \\
11b                       & 3                  & \textbf{73}              & \textbf{73}              & \textbf{73}          & \textbf{73}           & \textbf{73}         & \textbf{73}     & 73.0    \\
12a                       & 3                  & \textbf{77}              & \textbf{77}              & \textbf{77}          & \textbf{77}           & \textbf{77}         & \textbf{77}     & 77.0    \\
12b                       & 3                  & \textbf{983}             & \textbf{983}             & \textbf{983}         & \textbf{983}          & \textbf{983}        & \textbf{983}    & 983.0   \\
16                        & 3                  & \textbf{94}              & \textbf{94}              & \textbf{94}          & \textbf{94}           & \textbf{94}         & \textbf{94}     & 94.0\vspace{0.5em}   \\
128                       & 10                 & 4872                     & 4660                     & 4450                 & 2980                  & 2921                & \textbf{2547}   & 2583.7  \\
                          & 15                 & 3819                     & 3958                     & 3665                 & 2305                  & 2406                & \textbf{2053}   & 2072.7  \\
                          & 30                 & 3456                     & 3811                     & 3494                 & 1980                  & 2064                & \textbf{1859}   & 1896.3  \\ \bottomrule
\end{tabular}
\end{table*}

\subsection{Various variants of ITSHA}
This subsection presents various variant algorithms of ITSHA for comparison and analysis. The variant algorithms are as follows:

\begin{itemize}
\item \textbf{ITSHA-2opt}: A variant of ITSHA using only the operator 2-\emph{opt} in VNS.
\item \textbf{ITSHA-Insert}: A variant of ITSHA using only the operator \emph{Insert} in VNS.
\item \textbf{ITSHA-Swap}: A variant of ITSHA using only the operator \emph{Swap} in VNS.
\item \textbf{ITSHA-Fix$CS$}: A variant of ITSHA without adjusting the candidate sets at the end of each iteration.
\item \textbf{ITSHA-Zero$A_t$}: A variant of ITSHA without adjusting the solutions during the improvement stage.
\item \textbf{ITSHA-NoFCM}: A variant of ITSHA without the fuzzy clustering process during the initialization stage.
\textcolor{black}{\item \textbf{ITSHA-inter-intra}: A variant of ITSHA that restricts that the operator 2-\emph{opt} can only perform inter-tour improvements, and the operators \emph{Insert} and \emph{Swap} can only perform intra-tour improvements.}
\textcolor{black}{\item \textbf{ITSHA-Operator1}: A variant of ITSHA that replaces the proposed operators with the operators in \cite{Karabulut2021}.}
\textcolor{black}{\item \textbf{ITSHA-Operator2}: A variant of ITSHA that replaces the proposed operators with the operators in \cite{Wang2017}. Note that the operators in \cite{Soylu2015} are contained by those in \cite{Wang2017}.}
\item \textbf{ITSHA-$k$}: A variant of ITSHA that $C_{max}=k$ (we tested $k=5, 20, 50, n$ in experiments), note that ITSHA is equal to ITSHA-10, and ITSHA-$n$ indicates that the candidate set of each city contains all the other cities. 
\end{itemize}

We tested our ITSHA algorithm on all the instances in the four sets described in Section \ref{sec_instances}, \textcolor{black}{but mainly compared ITSHA with its variants on the most popular and widely used $m$TSP instances of \textit{Sets} \textit{\uppercase\expandafter{\romannumeral1}} and \textit{\uppercase\expandafter{\romannumeral2}} to evaluate the effectiveness of the components in ITSHA.}

\subsection{Comparison on ITSHA and the baselines}
Finally, we compare our ITSHA algorithm with the baseline algorithms. Tables \ref{table_set1}, \ref{table_set2}, \ref{table_set3}, and \ref{table_set4} show the results of ITSHA and the baseline algorithms in solving the instances of \textit{Sets \uppercase\expandafter{\romannumeral1}}, \textit{\uppercase\expandafter{\romannumeral2}}, \textit{\uppercase\expandafter{\romannumeral3}}, and \textit{\uppercase\expandafter{\romannumeral4}}, respectively. \textcolor{black}{In Tables \ref{table_set1}, \ref{table_set2}, and \ref{table_set4}, we provide the best and average solutions of ITSHA, and the best solutions of the other algorithms. Table \ref{table_set3} compares the best and the average solutions of ITSHA and GAL \cite{Lo2018}.}

%\textcolor{red}{!!!!!!!!!!!!!!!!!!!!}  In Tables \ref{table_set1}, \ref{table_set2} and \ref{table_set4}, we compare the best solution and the average solution of 10 runs of ITSHA, and the results of the other algorithms are the best solutions of them. And we compare the best solution and the average solution of ITSHA and GAL in Table \ref{table_set3}.

\begin{table*}[ht]
\caption{Comparison of ITSHA and the baseline algorithms in solving the \emph{minsum} and \emph{minmax} $m$TSP on the instances of \textit{Set \uppercase\expandafter{\romannumeral2}}. Best results appear in bold.}
\label{table_set2}
\begin{tabular}{lrrrrrrrrr} \toprule
\multirow{2}{*}{Instance} & \multirow{2}{*}{$m$} & \multirow{2}{*}{ABC(FC)} & \multirow{2}{*}{ABC(VC)} & \multirow{2}{*}{IWO} & \multirow{2}{*}{GVNS} & \multirow{2}{*}{ACO} & \multirow{2}{*}{ES} & \multicolumn{2}{r}{ITSHA} \\ \cline{9-10} 
                          &                    &                          &                          &                      &                       &                           &                     & Best            & Average \\ \hline
\multicolumn{10}{l}{\emph{minsum}}                                                                                                                                                                                                       \\ \hline
eil51    & 3                  & 446                      & 446                      & 446                  & 446                   & 446                       & 446                 & \textbf{443}    & 443.0   \\
                          & 5                  & 475                      & 472                      & 472                  & 472                   & 472                       & 472                 & \textbf{468}    & 468.0   \\
                          & 10                 & 580                      & 580                      & 581                  & 580                   & 580                       & 580                 & \textbf{577}    & 577.0\vspace{0.5em}   \\
kroD100  & 3                  & 21798                    & 21798                    & 21798                & 21879                 & 21798                     & 21798               & \textbf{21796}  & 21796.0  \\
                          & 5                  & 23238                    & 23182                    & 23294                & 23175                 & 23296                     & 23175               & \textbf{23173}  & 23173.0 \\
                          & 10                 & 27023                    & 26961                    & 26961                & 27008                 & 26966                     & 26927               & \textbf{26925}  & 26925.0 \\
                          & 20                 & 39509                    & 38333                    & \textbf{38245}       & 38326                 & \textbf{38245}            & \textbf{38245}      & 38248           & 38248.0\vspace{0.5em} \\
mTSP150  & 3                  & 38276                    & 38066                    & 37957                & 38430                 & 37958                     & 38072               & \textbf{37914}  & 37947.9 \\
                          & 5                  & 39309                    & 38979                    & \textbf{38714}       & 39171                 & 38729                     & 38907               & 38718           & 38782.7 \\
                          & 10                 & 43038                    & 42441                    & 42234                & 42703                 & 42234                     & \textbf{42203}      & 42206           & 42255.7 \\
                          & 20                 & 54279                    & 53603                    & 53475                & 53576                 & 53475                     & 53343               & \textbf{53310}  & 53386.9 \\
                          & 30                 & 69048                    & 68865                    & 68541                & 68558                 & 68541                     & 68606               & \textbf{68445}  & 68527.3 \\ \hline
\multicolumn{10}{l}{\emph{minmax}}                                                                                                                                                                                                       \\ \hline
eil51    & 3                  & 160                      & 160                      & 160                  & 160                   & 160                       & 160                 & \textbf{159}    & 159.0   \\
                          & 5                  & \textbf{118}             & \textbf{118}             & \textbf{118}         & \textbf{118}          & \textbf{118}              & \textbf{118}        & \textbf{118}    & 118.0   \\
                          & 10                 & \textbf{112}             & \textbf{112}             & \textbf{112}         & \textbf{112}          & \textbf{112}              & \textbf{112}        & \textbf{112}    & 112.0\vspace{0.5em}   \\
kroD100  & 3                  & 8577                     & 8509                     & 8509                 & 8509                  & 8511                      & 8509                & \textbf{8507}   & 8507.0  \\
                          & 5                  & 6785                     & 6768                     & 6767                 & 6767                  & 6845                      & \textbf{6766}       & 6770            & 6774.1  \\
                          & 10                 & \textbf{6358}            & \textbf{6358}            & \textbf{6358}        & \textbf{6358}         & \textbf{6358}             & \textbf{6358}       & \textbf{6358}   & 6358.0  \\
                          & 20                 & \textbf{6358}            & \textbf{6358}            & \textbf{6358}        & \textbf{6358}         & \textbf{6358}             & \textbf{6358}       & \textbf{6358}   & 6358.0\vspace{0.5em}  \\
mTSP150  & 3                  & 13896                    & 13461                    & 13168                & 13376                 & 13169                     & 13151               & \textbf{13084}  & 13236.5 \\
                          & 5                  & 8889                     & 8678                     & 8479                 & 8467                  & 8467                      & 8466                & \textbf{8465}   & 8543.6  \\
                          & 10                 & 5803                     & 5728                     & 5594                 & 5674                  & 5565                      & \textbf{5557}       & \textbf{5557}   & 5604.0  \\
                          & 20                 & \textbf{5246}            & \textbf{5246}            & \textbf{5246}        & \textbf{5246}         & \textbf{5246}             & \textbf{5246}       & \textbf{5246}   & 5246.0  \\
                          & 30                 & \textbf{5246}            & \textbf{5246}            & \textbf{5246}        & \textbf{5246}         & 5247                      & \textbf{5246}       & \textbf{5246}   & 5246.0 \\ \bottomrule
\end{tabular}
\end{table*}

\begin{table*}[ht]
\caption{Comparison of ITSHA and the baseline algorithms in solving the \emph{minsum} $m$TSP on the instances of \textit{Set \uppercase\expandafter{\romannumeral3}}. Column $p$ indicates the maximum number of cities that can be visited by any salesman. Best results appear in bold.}
\label{table_set3}
\begin{tabular}{lrrrrrrrr} \toprule
\multirow{2}{*}{Instance} & \multirow{2}{*}{$p$} & \multirow{2}{*}{$m$} &  & \multicolumn{2}{r}{GAL}    &  & \multicolumn{2}{r}{ITSHA}   \\ \cline{5-6} \cline{8-9} 
                          &                    &                    &  & Best            & Average  &  & Best            & Average  \\ \hline
pr76                      & 20                 & 5                  &  & \textbf{152278} & 166138.0 &  & 152972          & 153237.0 \\
                          &                    & 10                 &  & 177806          & 182381.0 &  & \textbf{175676} & 175764.0 \\
                          &                    & 15                 &  & 218901          & 223927.0 &  & \textbf{216294} & 216294.0\vspace{0.5em} \\
pr152                     & 40                 & 5                  &  & 116620          & 131674.0 &  & \textbf{113887} & 114023.3 \\
                          &                    & 10                 &  & 132917          & 141993.0 &  & \textbf{122732} & 122863.8 \\
                          &                    & 15                 &  & 154249          & 164741.0 &  & \textbf{143631} & 143702.4\vspace{0.5em} \\
pr226                     & 50                 & 5                  &  & 148040          & 156629.0 &  & \textbf{144560} & 147462.5 \\
                          &                    & 10                 &  & 167782          & 171338.0 &  & \textbf{154188} & 158838.4 \\
                          &                    & 15                 &  & 180431          & 188489.0 &  & \textbf{170320} & 174099.2\vspace{0.5em} \\
pr299                     & 70                 & 5                  &  & 73177           & 77676.0  &  & \textbf{69329}  & 70220.9  \\
                          &                    & 10                 &  & 75450           & 78999.0  &  & \textbf{71803}  & 72611.3  \\
                          &                    & 15                 &  & 84266           & 87490.0  &  & \textbf{79991}  & 80740.2\vspace{0.5em}  \\
pr439                     & 100                & 5                  &  & 141180          & 147389.0 &  & \textbf{133197} & 133930.8 \\
                          &                    & 10                 &  & 144527          & 151392.0 &  & \textbf{135930} & 137000.1 \\
                          &                    & 15                 &  & 149649          & 155512.0 &  & \textbf{140948} & 141707.5\vspace{0.5em} \\
pr1002                    & 220                & 5                  &  & 332652          & 338580.0 &  & \textbf{303142} & 308753.2 \\
                          &                    & 10                 &  & 347126          & 360284.0 &  & \textbf{317752} & 320886.5 \\
                          &                    & 15                 &  & 379677          & 383360.0 &  & \textbf{339448} & 343863.1 \\ \bottomrule
\end{tabular}
\end{table*}

\begin{table*}[ht]
\caption{Comparison of ITSHA and the baseline algorithms in solving the \emph{minmax} $m$TSP on the instances of \textit{Set \uppercase\expandafter{\romannumeral4}}. Best results appear in bold.}
\label{table_set4}
\begin{tabular}{lrrrrrrrrr} \toprule
\multirow{2}{*}{Instance} & \multirow{2}{*}{$m$} & \multirow{2}{*}{ABC(FC)} & \multirow{2}{*}{ABC(VC)} & \multirow{2}{*}{IWO} & \multirow{2}{*}{GVNS} & \multirow{2}{*}{MASVND} & \multirow{2}{*}{ES} & \multicolumn{2}{r}{ITSHA}    \\ \cline{9-10} 
                          &                    &                          &                          &                      &                       &                         &                     & Best              & Average  \\ \hline
ch150                     & 3                  & 2454.40                  & 2437.04                  & 2413.24              & 2427.77               & 2429.49                 & 2407.59             & \textbf{2405.94}  & 2435.25  \\
                          & 5                  & 1797.32                  & 1764.65                  & 1752.11              & 1830.50               & 1758.08                 & 1741.61             & \textbf{1740.63}  & 1765.62  \\
                          & 10                 & 1563.39                  & 1557.94                  & 1554.64              & 1554.64               & 1554.64                 & 1554.64             & \textbf{1554.33}  & 1554.33  \\
                          & 20                 & 1554.64                  & 1554.64                  & 1554.64              & 1554.64               & 1554.64                 & 1554.64             & \textbf{1554.33}  & 1554.33\vspace{0.5em}  \\
kroA200                   & 3                  & 10976.60                 & 10933.11                 & 10814.18             & 10898.96              & 10831.66                & 10768.10            & \textbf{10760.69} & 10892.27 \\
                          & 5                  & 7795.41                  & 7595.41                  & 7493.24              & 7836.21               & \textbf{7415.54}        & 7572.32             & 7470.78           & 7547.11  \\
                          & 10                 & 6291.01                  & 6294.24                  & 6237.00              & \textbf{6223.22}      & \textbf{6223.22}        & \textbf{6223.22}    & \textbf{6223.22}  & 6223.22  \\
                          & 20                 & \textbf{6223.22}         & \textbf{6223.22}         & \textbf{6223.22}     & \textbf{6223.22}      & \textbf{6223.22}        & \textbf{6223.22}    & \textbf{6223.22}  & 6223.22\vspace{0.5em} \\
lin318                    & 3                  & 17062.22                 & 16707.02                 & 16200.21             & 16861.99              & 16206.25                & 16273.80            & \textbf{15918.24} & 16237.07 \\
                          & 5                  & 12449.06                 & 12088.64                 & 11730.03             & 12210.40              & 11752.41                & 11604.20            & \textbf{11548.44} & 11811.14 \\
                          & 10                 & 10061.30                 & 9983.23                  & 9845.72              & 9826.77               & \textbf{9731.17}        & \textbf{9731.17}    & \textbf{9731.17}  & 9731.17  \\
                          & 20                 & \textbf{9731.17}         & \textbf{9731.17}         & \textbf{9731.17}     & \textbf{9731.17}      & \textbf{9731.17}        & \textbf{9731.17}    & \textbf{9731.17}  & 9731.17\vspace{0.5em} \\
att532                    & 3                  & 35138.50                 & 34401.24                 & 32988.99             & 36395.54              & 32403.10                & 33597.40            & \textbf{32223.24} & 32882.79 \\
                          & 5                  & 25033.97                 & 24564.33                 & 23519.68             & 24866.28              & 22619.66                & 23089.70            & \textbf{22372.68} & 22867.24 \\
                          & 10                 & 19949.41                 & 19584.52                 & 19136.52             & 19278.83              & 18390.46                & \textbf{18059.70}   & 18091.15          & 18313.77 \\
                          & 20                 & 18332.84                 & 18156.62                 & 17850.80             & 17822.23              & \textbf{17641.16}       & 17641.20            & \textbf{17641.16} & 17700.95\vspace{0.5em} \\
rat783                    & 3                  & 3622.79                  & 3530.31                  & 3457.97              & 3518.08               & 3279.16                 & 3369.40             & \textbf{3158.34}  & 3227.53  \\
                          & 5                  & 2413.85                  & 2317.66                  & 2273.80              & 2325.47               & 2092.77                 & 2127.99             & \textbf{2024.27}  & 2077.57  \\
                          & 10                 & 1626.69                  & 1587.43                  & 1542.05              & 1515.03               & 1432.34                 & \textbf{1360.89}    & 1367.98           & 1393.77  \\
                          & 20                 & 1375.16                  & 1350.96                  & 1311.30              & 1578.87               & 1260.88                 & \textbf{1231.69}    & \textbf{1231.69}  & 1235.00\vspace{0.5em} \\
pcb1173                   & 3                  & 24748.31                 & 24384.17                 & 24008.47             & 24988.00              & 22443.22                & 22601.70            & \textbf{20292.61} & 20675.14 \\
                          & 5                  & 16590.98                 & 16222.91                 & 16057.19             & 15494.12              & 14557.30                & 14099.50            & \textbf{12952.97} & 13227.20 \\
                          & 10                 & 10965.66                 & 10652.46                 & 10517.94             & 10386.45              & 9222.92                 & 8160.25             & \textbf{7864.11}  & 8000.58  \\
                          & 20                 & 8373.09                  & 8228.66                  & 8063.17              & 8311.38               & 7063.23                 & 6549.14             & \textbf{6528.86}  & 6584.69 \\ \bottomrule
\end{tabular}
\end{table*}

From the results in Tables \ref{table_set1}, \ref{table_set2}, \ref{table_set3}, and \ref{table_set4}, we observe that our ITSHA algorithm significantly outperforms other state-of-the-art heuristic algorithms in solving both the \emph{minsum} and \emph{minmax} $m$TSP. \textcolor{black}{Specifically, ITSHA yields 6/3 new best-known solutions for all the 8 instances of \textit{Set \uppercase\expandafter{\romannumeral1}} with the \emph{minsum}/\emph{minmax} objective, 9/4 new best-known solutions for all the 12 instances of \textit{Set \uppercase\expandafter{\romannumeral2}} with the \emph{minsum}/\emph{minmax} objective, 17 new best-known solutions for all the 18 \emph{minsum} $m$TSP instances of \textit{Set \uppercase\expandafter{\romannumeral3}}, and 15 new best-known solutions for all the 24 \emph{minmax} $m$TSP instances of \textit{Set \uppercase\expandafter{\romannumeral4}}. In summary, for all the 38 tested \emph{minsum} $m$TSP instances, ITSHA can yield better results than the best-known solutions in the literature on 32 instances. For all the 44 tested \emph{minmax} $m$TSP instances, ITSHA can yield better results than the best-known solutions in the literature on 22 instances.} %Moreover, the improvement of ITSHA over other state-of-the-art heuristic algorithms is more obvious for large instances. 
The results demonstrate that our proposed ITSHA algorithm is powerful and effective in solving the $m$TSP on both the objectives.

\subsection{Comparison on local search operators}
\begin{figure*}[t]
\centering
\subfigure[Comparison on ITSHA with different local search operators in solving the \emph{minsum} $m$TSP]{
\includegraphics[width=1.5\columnwidth]{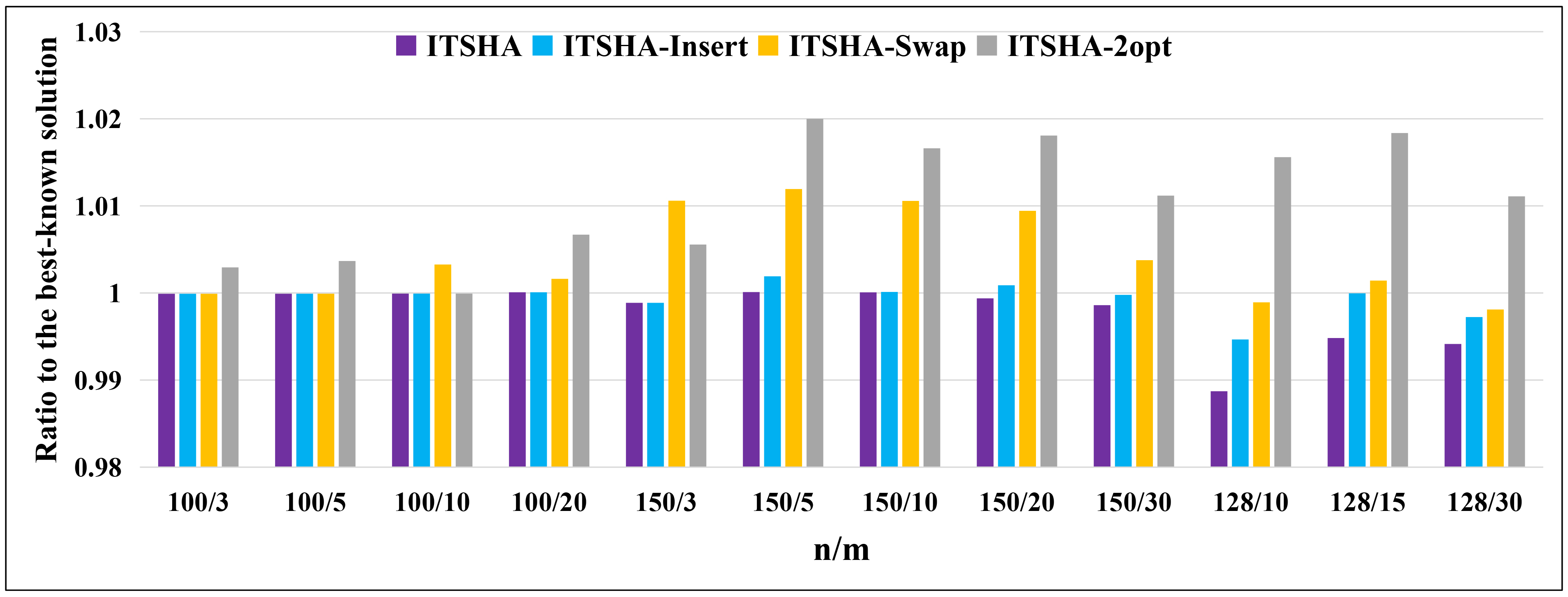}  
\label{fig_operators-minsum}}\hspace{1em}
\subfigure[Comparison on ITSHA with different local search operators in solving the \emph{minmax} $m$TSP]{
\includegraphics[width=1.5\columnwidth]{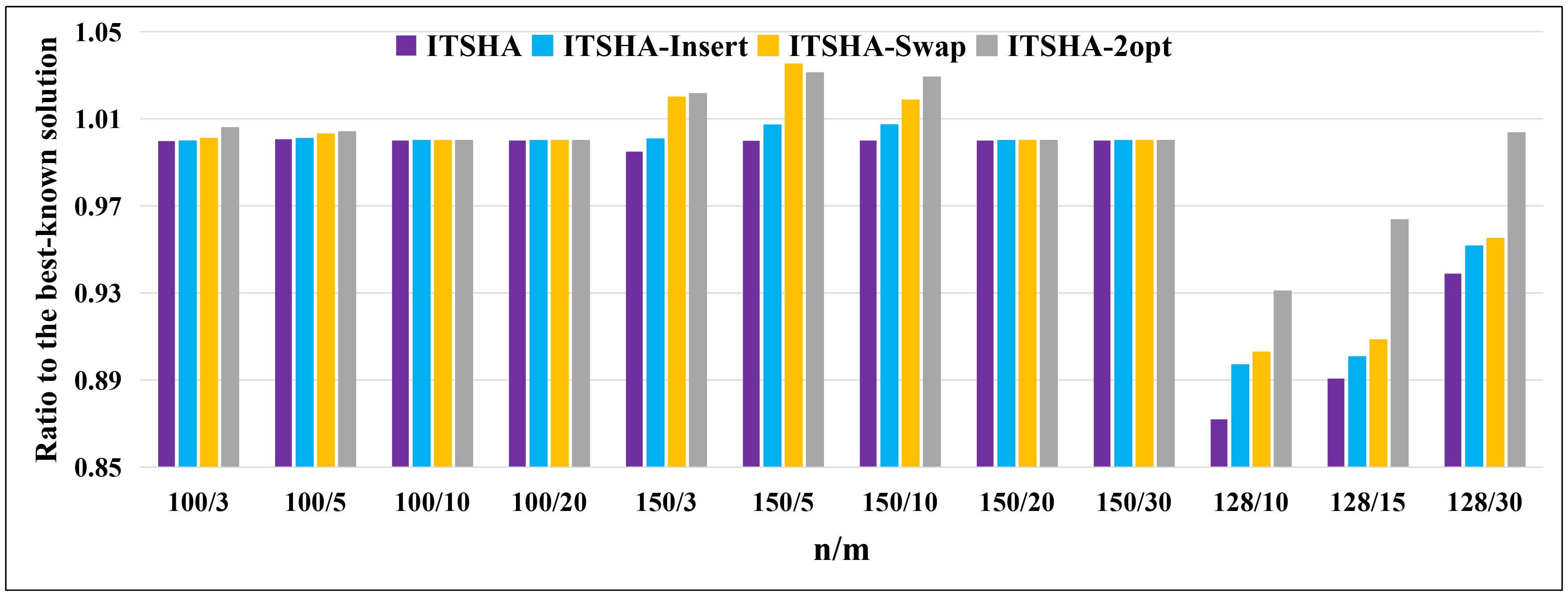} 
\label{fig_operators-minmax}}
\caption{Evaluation of ITSHA with different local search operators in solving the \emph{minsum} and \emph{minmax} $m$TSP.}
\label{fig_operators}
\end{figure*}

We first compare the performance of the three local search operators used in ITSHA, including 2-\emph{opt}, \emph{Insert}, and \emph{Swap}. Figure \ref{fig_operators} shows the results of ITSHA, ITSHA-2opt, ITSHA-Insert, and ITSHA-Swap in solving the instances in \textit{Sets} \textit{\uppercase\expandafter{\romannumeral1}} and \textit{\uppercase\expandafter{\romannumeral2}} except the instances with $n=11,12,16,51$ (since they are too simple to distinguish the performance of the algorithms). The results are expressed by the ratio of the best solutions obtained in 10 runs of the four algorithms to the best-known solutions in the literature \cite{Venkatesh2015,Soylu2015,Lu2019,Karabulut2021}.

From the results in Figure \ref{fig_operators}, we can see that:

(1) The order of the three local search operators with decreasing performance is: \emph{Insert}, \emph{Swap}, and 2-\emph{opt}. Therefore, we order \emph{Insert} first, then \emph{Swap}, and finally 2-\emph{opt} in the VNS process of ITSHA.

(2) The performance of ITSHA is better than that of the other three variants, indicating that the VNS method can make use of the complimentary of the three local search operators in searching for better solutions and improve the performance.

\textcolor{black}{(3) Our algorithm shows better performance for solving the instance \emph{128} than the other instances from the TSPLIB (i.e., \emph{kroD100} and \emph{mTSP150}). This might be because the structure of instance \emph{128} is different from that of the instances from the TSPLIB, which is difficult for the baseline algorithms. ITSHA can still solve the instance \emph{128} well, indicating the good robustness of our method for solving instances from various datasets.}

\begin{figure*}[t]
\centering
\subfigure[Comparison on different size of candidate sets in solving the \emph{minsum} $m$TSP]{
\includegraphics[width=1.5\columnwidth]{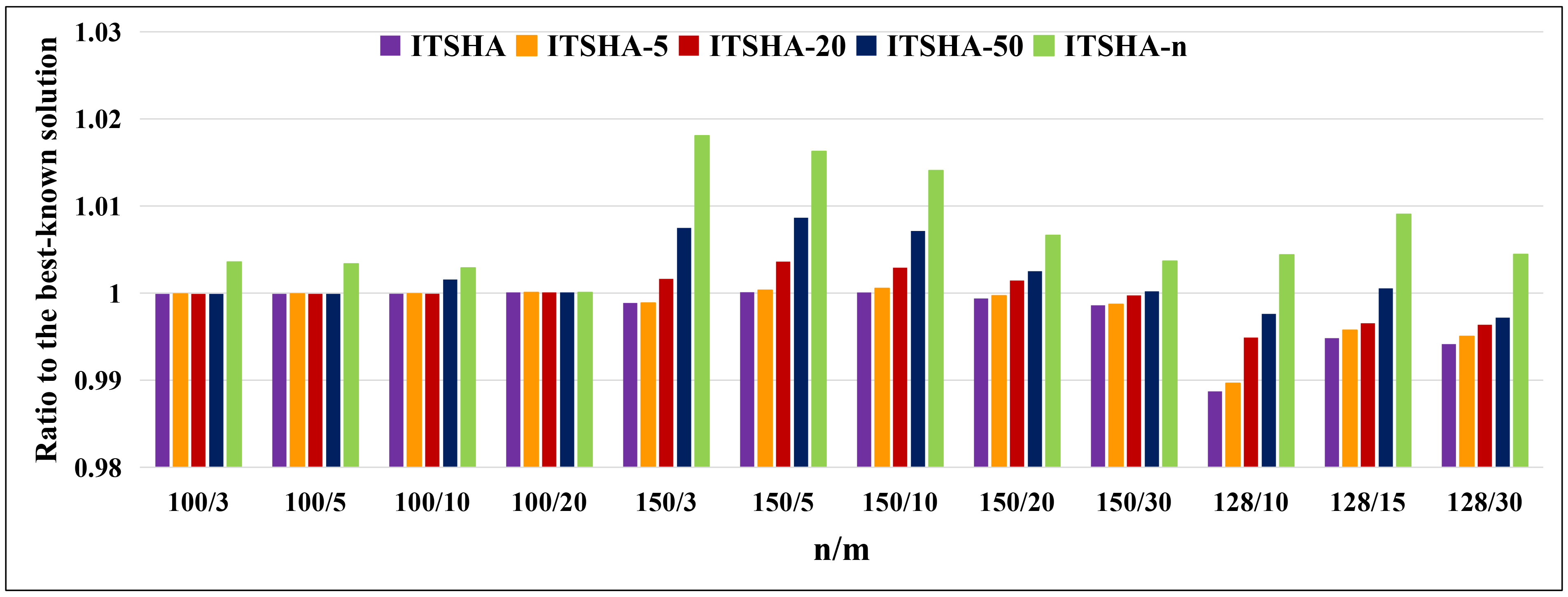}  
\label{fig_ITSHA-k-minsum}}\hspace{1em}
\subfigure[Comparison on different size of candidate sets in solving the \emph{minmax} $m$TSP]{
\includegraphics[width=1.5\columnwidth]{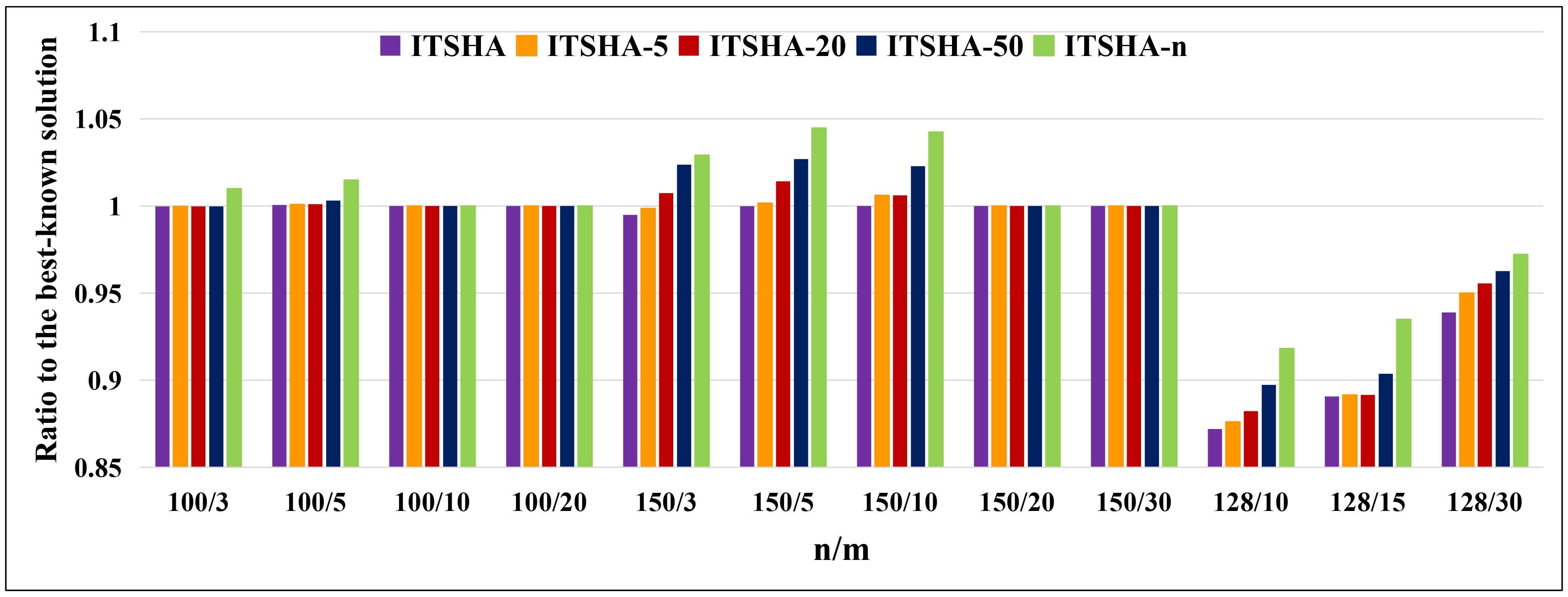} 
\label{fig_ITSHA-k-minmax}}
\caption{Evaluation of ITSHA with different sizes of candidate sets in solving the \emph{minsum} and \emph{minmax} $m$TSP.}
\label{fig_ITSHA-k}
\end{figure*}

\subsection{Comparison on different size of candidate sets}
We further compare the ITSHA algorithm with different sizes of candidate sets. Figure \ref{fig_ITSHA-k} shows the results of ITSHA, ITSHA-5, ITSHA-20, ITSHA-50, and ITSHA-$n$ in solving the same 12 instances in Figure \ref{fig_operators}. The results are expressed by the ratio of the best solution obtained in 10 runs by these five algorithms to the best-known solutions in the literature \cite{Venkatesh2015,Soylu2015,Lu2019,Karabulut2021}.

From the results in Figure \ref{fig_ITSHA-k}, we can see that:

(1) ITSHA with $C_{max}=10$ (the default value) is the best-performing algorithm among the five algorithms in Figure \ref{fig_ITSHA-k}. The decrease of $C_{max}$ may reduce the search ability of the algorithm, and the increase of $C_{max}$ may reduce the efficiency of the algorithm.

(2) The performance of ITSHA-$n$ is much worse than the other four algorithms in Figure \ref{fig_ITSHA-k}, demonstrating that the candidate sets can significantly improve the performance of the VNS process in ITSHA. \textcolor{black}{The results also indicate that the search neighborhoods in ITSHA are much more efficient than those in \cite{Soylu2015,Karabulut2021,Wang2017}, since they do not use the candidate sets to refine the search region, but traverse all the possible operators.}

\textcolor{black}{(3) The results demonstrate again that ITSHA shows excellent performance in solving the instance \emph{128}, as ITSHA-$k$ with $k=(5,20)$ can yield solutions better than the best-known solutions of instance \emph{128} with the \emph{minsum} objective, and ITSHA-$k$ with $k=(5,20,50,n)$ can yield solutions better than the best-known solutions of instance \emph{128} with the \emph{minmax} objective.}

\begin{table*}[t]
\caption{Comparison of ITSHA with its variants ITSHA-Fix$CS$, ITSHA-Zero$A_t$, and ITSHA-NoFCM.}
\label{table_ablation}
\scalebox{0.9}{\begin{tabular}{lrrrrrrrrrrrrrrr} \toprule
Instance      &    & \multicolumn{4}{r}{kroD100}   &  & \multicolumn{5}{r}{mTSP150}           &  & \multicolumn{3}{r}{128} \\ \cline{3-6} \cline{8-12} \cline{14-16} 
$m$       &          & 3     & 5     & 10    & 20    &  & 3     & 5     & 10    & 20    & 30    &  & 10     & 15     & 30    \\ \hline
\multicolumn{16}{l}{\emph{minsum}}                                                                                               \\ \hline
ITSHA-Fix$CS$   &    & 21796 & 23173 & 26925 & 38248 &  & 37914 & 38720 & 42217 & 53411 & 68552 &  & 21144  & 23882  & 36706 \\
ITSHA-Zero$A_t$  &    & 21796 & 23173 & 26925 & 38248 &  & 37914 & 38778 & 42206 & 53366 & 68453 &  & 21123  & 23884  & 36668 \\
ITSHA-NoFCM   &    & 21796 & 23173 & 26925 & 38248 &  & 37947 & 38751 & 42208 & 53310 & 68459 &  & 21145  & 23862  & 36681 \\
%ITSHA-inter-intra & & 21796 & 23173 & 26961 & 38248 &  & 38001 & 39046 & 42572 & 53719 & 68754 &  & 21202  & 23966  & 36791 \\
ITSHA      &       & 21796 & 23173 & 26925 & 38248 &  & 37914 & 38718 & 42206 & 53310 & 68445 &  & 21113  & 23838  & 36655 \\ \hline
\multicolumn{16}{l}{\emph{minmax}}                                                                                               \\ \hline
ITSHA-Fix$CS$   &    & 8507  & 6773  & 6358  & 6358  &  & 13177 & 8493  & 5591  & 5246  & 5246  &  & 2582   & 2055   & 1884  \\
ITSHA-Zero$A_t$   &   & 8507  & 6770  & 6358  & 6358  &  & 13131 & 8507  & 5591  & 5246  & 5246  &  & 2556   & 2058   & 1874  \\
ITSHA-NoFCM   &    & 8507  & 6772  & 6358  & 6358  &  & 13168 & 8481  & 5593  & 5246  & 5246  &  & 2563   & 2068   & 1870  \\
%ITSHA-inter-intra & & 8507  & 6772  & 6358  & 6358  &  & 13380 & 8616  & 5640  & 5246  & 5246  &  & 2618   & 2103   & 1927  \\
ITSHA      &       & 8507  & 6770  & 6358  & 6358  &  & 13084 & 8465  & 5557  & 5246  & 5246  &  & 2547   & 2053   & 1859  \\ \bottomrule
\end{tabular}}
\end{table*}

\begin{table*}[t]
\caption{Comparison of ITSHA with its variants ITSHA-inter-intra, ITSHA-Operator1, and ITSHA-Operator2.}
\label{table_operators}
\scalebox{0.9}{\begin{tabular}{lrrrrrrrrrrrrrrr} \toprule
Instance          &  & \multicolumn{4}{r}{kroD100}   &  & \multicolumn{5}{r}{mTSP150}           &  & \multicolumn{3}{r}{128} \\ \cline{3-6} \cline{8-12} \cline{14-16} 
$m$                 &  & 3     & 5     & 10    & 20    &  & 3     & 5     & 10    & 20    & 30    &  & 10     & 15     & 30    \\ \hline
\multicolumn{16}{l}{\emph{minsum}}                                                                                                  \\ \hline
ITSHA-inter-intra &  & 21796 & 23173 & 26961 & 38248 &  & 38001 & 39046 & 42572 & 53719 & 68754 &  & 21202  & 23966  & 36791 \\
ITSHA-Operator1   &  & 22033 & 23404 & 27504 & 39228 &  & 39034 & 40279 & 43832 & 55828 & 70617 &  & 29934  & 33495  & 45385 \\
ITSHA-Operator2   &  & 21796 & 23173 & 26925 & 38248 &  & 38037 & 39129 & 42640 & 53752 & 68926 &  & 23447  & 26187  & 38515 \\
ITSHA             &  & 21796 & 23173 & 26925 & 38248 &  & 37914 & 38718 & 42206 & 53310 & 68445 &  & 21113  & 23838  & 36655 \\ \hline
\multicolumn{16}{l}{\emph{minmax}}                                                                                                  \\ \hline
ITSHA-inter-intra &  & 8507  & 6772  & 6358  & 6358  &  & 13380 & 8616  & 5640  & 5246  & 5246  &  & 2618   & 2103   & 1927  \\
ITSHA-Operator1   &  & 8677  & 6806  & 6358  & 6358  &  & 13883 & 8790  & 5696  & 5246  & 5246  &  & 3549   & 2701   & 2208  \\
ITSHA-Operator2   &  & 8507  & 6772  & 6358  & 6358  &  & 13314 & 8568  & 5595  & 5246  & 5246  &  & 2925   & 2305   & 1969  \\
ITSHA             &  & 8507  & 6770  & 6358  & 6358  &  & 13084 & 8465  & 5557  & 5246  & 5246  &  & 2547   & 2053   & 1859 \\ \bottomrule
\end{tabular}}
\end{table*}

\subsection{Analyses on local optima escaping approaches}
We then analyze the effectiveness of the approach of adjusting candidate sets, the solution adjustment process, the fuzzy clustering algorithm in our ITSHA algorithm.%, \textcolor{black}{as well as the mechanism that all the neighborhoods can be used to perform inter-tour or intra-tour improvements in our ITSHA algorithm.} 
Table \ref{table_ablation} shows the results of ITSHA-Fix$CS$, ITSHA-Zero$A_t$, ITSHA-NoFCM, and ITSHA in solving the same 12 instances in Figure \ref{fig_operators}. The results are expressed by the best solutions obtained in 10 runs by these four algorithms.

From the results we can observe that:

\textcolor{black}{(1) ITSHA outperforms ITSHA-Fix$CS$, indicating that the method of adjusting the candidate sets can improve the performance of ITSHA by helping the algorithm escape from the local optima.}

\textcolor{black}{(2) ITSHA outperforms ITSHA-Zero$A_t$, indicating that the solution adjustment process can also improve the performance. Our ITSHA algorithm does not allow the current solution to be worse than the previous one. Adjusting the solution in each iteration can improve the flexibility and search ability of the algorithm.}

\textcolor{black}{(3) ITSHA outperforms ITSHA-NoFCM, indicating that the fuzzy clustering process can improve the algorithm by providing higher-quality initial solutions, since the clustering algorithm can allocate the cities with close positions to one salesman.}

%\textcolor{black}{(4) ITSHA significantly outperforms ITSHA-inter-intra, demonstrating that our search neighborhoods that can perform both inter-tour and intra-tour improvements are effective. The results also show that our search neighborhoods are significantly better than those in \cite{Soylu2015} and \cite{Wang2017}, which only apply the 2-\emph{opt} operator to perform inter-tour improvements and other operators to perform intra-tour improvements.} 

%Figure \ref{fig_ablation} shows the results of ITSHA-Fix$CS$, ITSHA-Zero$A_t$ and ITSHA-NoFCM algorithms in solving the same 9 instances in Figure \ref{fig_operators}. The results are expressed by the ratio of the best solutions obtained in 10 runs by these three algorithms to the best solution obtained in 10 runs of ITSHA.

%From the results in Figure \ref{fig_operators}, we observe that the performance of the three variants are all worse than that of ITSHA in solving both of the \emph{minsum} and the \emph{minmax} $m$TSP, indicating that the method of adjusting candidate sets, the solution adjustment process and the fuzzy clustering process can all enhance the performance of ITSHA by helping the algorithm to escape from the local optima and find better solutions.

\textcolor{black}{\subsection{Analyze the superiority of proposed operators}}

\textcolor{black}{In order to demonstrate the advantages of the proposed operators, we compared ITSHA with its variants, including ITSHA-inter-intra, ITSHA-Operator1, and ITSHA-Operator2, for solving the 12 instances in Figure \ref{fig_operators}. Table \ref{table_operators} compares the best solutions obtained in 10 runs by these four algorithms.}

\textcolor{black}{From the results we can see that:}

\textcolor{black}{(1) ITSHA significantly outperforms ITSHA-inter-intra, demonstrating that our search neighborhoods that can perform both inter-tour and intra-tour improvements are effective. Such a mechanism is one of the factors that explain the success of our proposed operators.}

\textcolor{black}{(2) ITSHA significantly outperforms ITSHA-Operator1 and ITSHA-Operator2, indicating that our proposed operators are much better than the operators in \cite{Soylu2015,Karabulut2021,Wang2017}. The reasons why our operators show much better performance are as follows. First, we apply the candidate sets to refine the search region. Second, the operators \emph{Insert} and \emph{Swap} can move a sequence of cities, which can help the algorithm find high-quality solutions. Third, our operators can perform both inter-tour and intra-tour improvements.}

\textcolor{black}{(3) ITSHA-Operator1 is the worst among the four compared algorithms, because the operators in the ES algorithm \cite{Karabulut2021} are too simple and ES mainly depends on the evolution method to find high-quality solutions.}

%The results also show that our search neighborhoods are significantly better than those in \cite{Soylu2015} and \cite{Wang2017}, which only apply the 2-\emph{opt} operator to perform inter-tour improvements and other operators to perform intra-tour improvements.

\section{Conclusion}
\label{sec_Con}
This paper proposes an iterated two-stage heuristic algorithm, called ITSHA, for the \emph{minsum} and \emph{minmax} multiple Traveling Salesmen Problem ($m$TSP). Each iteration of ITSHA consists of an initialization stage and an improvement stage. The initialization stage containing the fuzzy clustering algorithm and a proposed random greedy function is used to generate high-quality and diverse initial solutions. The improvement stage mainly applies the variable neighborhood search (VNS) approach based on our proposed neighborhoods (2-\emph{opt}, \emph{Insert}, and \emph{Swap}) to improve the initial solution. 

Our proposed neighborhoods are effective and efficient, which benefit from the employment of the candidate sets. We further apply some approaches to help the local search algorithm escape from the local optima, for example, the fuzzy clustering algorithm and the random greedy function in the initialization stage, the solution adjustment during the improvement stage, and the method of adjusting candidate sets at the end of each iteration of ITSHA. 

In summary, the ITSHA algorithm benefits from the effective and efficient VNS local search and the approaches for escaping from the local optima. Experimental results on 38 \emph{minsum} $m$TSP and 44 \emph{minmax} $m$TSP benchmarks demonstrate that our proposed ITSHA algorithm significantly outperforms the state-of-the-art heuristic algorithms in solving both the \emph{minsum} and \emph{minmax} $m$TSP. Moreover, \textcolor{black}{the proposed search neighborhoods and the local optima escaping approaches could be applied to other combinatorial optimization problems, such as the TSP, VRP, and their various variants.}

%\appendix
%\section{My Appendix}
\section*{Declarations of interest}
None.

\balance
\bibliographystyle{unsrt}

\bibliography{ITSHA}


\end{document}